\documentclass[lettersize,journal]{IEEEtran}

\AtBeginDocument{%
  }

\usepackage{graphicx}
\usepackage{subcaption}
\usepackage{float}
\usepackage{subfig}
\usepackage{multirow}
\usepackage{pifont}
\usepackage{color}
\usepackage{amsmath}

\newcommand{\ie}{\emph{i.e.} }

\newcommand{\eg}{\emph{e.g.} }

\newcommand{\etal}{\emph{et al.} }

\newcommand\figref[1]{Fig.~\ref{#1}}

\newcommand\secref[1]{Sec.~\ref{#1}}

\newcommand{\sysname}{{\sf CrowdHMTware} }

\ifodd 1

\newcommand\rev[1]{\textcolor{black}{#1}}

\newcommand\lsc[1]{\textcolor{black}{#1}}

\newcommand{\discuss}[1]{{\color{black}}}
\else
\fi

\begin{document}
%\definecolor{mypurple}{RGB}{128,0,128}

\title{\huge CrowdHMTware: A Cross-level Co-adaptation Middleware for Context-aware Mobile DL Deployment}

\author{Sicong Liu,~\IEEEmembership{Member,~IEEE,}~Bin Guo*,~\IEEEmembership{Senior Member,~IEEE,}~Shiyan Luo,~Yuzhan Wang,~Hao Luo,~Cheng Fang,~Yuan Xu,~Ke Ma,~Yao Li,~Zhiwen Yu,~\IEEEmembership{Senior Member,~IEEE}
        % <-this % stops a space
% \thanks{This paper was produced by the IEEE Publication Technology Group. They are in Piscataway, NJ.}
\thanks{{*}Corresponding Author: Bin Guo (E-mail:guob@nwpu.edu.cn)}
% <-this % stops a space
% \thanks{Manuscript received October 15, 2024; revised October 15, 2024.}
}

\markboth{Journal of \LaTeX\ Class Files,~Vol.~14, No.~8, August~2021}%
{Shell \MakeLowercase{\textit{et al.}}: A Sample Article Using IEEEtran.cls for IEEE Journals}

\IEEEpubid{0000--0000/00\$00.00~\copyright~2021 IEEE}

\maketitle

\begin{abstract}
There are many deep learning (DL) powered mobile and wearable applications today continuously and unobtrusively sensing the ambient surroundings to enhance all aspects of human lives.
To enable robust and private mobile sensing, DL models are often deployed locally on resource-constrained mobile devices using techniques such as model compression or offloading.
However, existing methods, either front-end algorithm level (\ie DL model compression/partitioning) or back-end scheduling level (\ie operator/resource scheduling), cannot be locally online because
they require offline retraining to ensure accuracy or rely on manually pre-defined strategies, struggle with \textit{dynamic adaptability}.
The primary challenge lies in feeding back runtime performance from the \textit{back-end} level to the \textit{front-end} level optimization decision.
Moreover, the adaptive mobile DL model porting middleware with \textit{cross-level co-adaptation} is less explored, particularly in mobile environments with \textit{diversity} and \textit{dynamics}.
In response, we introduce CrowdHMTware, a dynamic context-adaptive DL model deployment middleware for heterogeneous mobile devices. 
It establishes an \textit{automated adaptation loop} between cross-level functional components, \ie elastic inference, scalable offloading, and model-adaptive engine, enhancing scalability and adaptability. 
Experiments with four typical tasks across 15 platforms and a real-world case study demonstrate that \sysname can effectively scale DL model, offloading, and engine actions across diverse platforms and tasks.
It hides run-time system issues from developers, reducing the required developer expertise.

\end{abstract}


\maketitle

\section{Introduction}

% 1DL 应用广泛，引出资源高效的必要性
There is a growing trend to integrate DL-powered intelligence into mobile and embedded devices across a wide range of applications~\cite{fang2024AdaShadow}, including Google Cloud Vision for object detection~\cite{cheng2023sfrnet}, Amazon SageMaker for semantic segmentation~\cite{thisanke2023semantic}, Google Coral for object tracking~\cite{chen2023seqtrack}, Apple Siri for natural language processing~\cite{bharadiya2023comprehensive}, and Microsoft Azure Personalizer for recommendation systems~\cite{liu2023pre}.
This trend is driven by the rise in mobile and embedded devices, which has increased the demand for processing rich sensor data locally to minimize bandwidth usage and reduce latency \cite{canziani2016analysis}. 
Moreover, mobile applications often handle sensitive personal data, such as in health assistance and security monitoring, raising significant privacy concerns \cite{zhou2019edge}. 
Despite these advancements, deploying resource-intensive DL models on mobile and embedded devices while maintaining accuracy and real-time performance remains a significant challenge.

Given these challenges, existing studies have explored various \textit{\textbf{front-end algorithm-level}} techniques such as \textit{handcrafted} or \textit{on-demand} model compression~\cite{liu2021adaspring} and partitioning~\cite{zeng2020coedge}, as well as \textit{\textbf{back-end scheduling-level}} optimizations like operator fusion~\cite{niu2021dnnfusion}, operator parallelism~\cite{zheng2022alpa}, and memory allocation/swapping~\cite{wang2022melon}. 
As the accuracy and responsiveness of DL models are bounded by resource availability, joint optimizing across algorithm and system scheduling levels with bi-directional feedback can improve runtime resource availability and thereby push performance \textit{boundaries} set by stand-alone level.
Previous DL compilers and {frameworks explore \textit{backend-level system scheduling optimization} after DL models are loaded to maximize resource utilization, data reuse, and minimize runtime overhead. 
However, they always rely on \textit{manual} effort at specific partial levels since the optimization criterion is a black box. 
For example, TensorflowXLA~\cite{tensorflowxla} introduces a linear algebra compiler engine specifically designed for the TensorFlow, while TVM~\cite{chen2018tvm} enhances DL \textit{operator} optimization at the intermediate computation graph level, enabling cross-framework support.

Despite existing efforts, the challenge of dynamically adapting the DL system \textit{extensively covering these levels} in real-time for mobile environments with inherent \textit{diversity} and \textit{dynamics} remains unresolved.
The unique characteristics of mobile contexts lie in their execution diversity and dynamics.
i) \textit{Diversity} includes variations in DL model architectures, mobile application-specified demands for accuracy, latency, and resource budgets, and the DL frameworks tailored to specific mobile devices. 
    It also includes heterogeneous hardware profiles, such as CPUs, GPUs, DSPs, and NPUs, each supporting unique operators and libraries.
ii) \textit{Dynamics} refer to changes in live data distribution, runtime resource availability (\ie light and heavy workloads), fluctuations in processor frequencies, varying inference requests, and unpredictable competition from other processes (\eg UI interactions).
Unlike the resource-rich cloud, this dynamism significantly impacts DL model performance on resource-limited mobile devices.
Subsequently, the diversity and dynamic nature in mobile contexts presents the following challenges for \textit{runtime cross-level optimization}.

\IEEEpubidadjcol

\noindent$\bullet$ \textbf{\textit{Challenge $\#$1}}: Performance demands (\eg accuracy, latency, memory usage) often interdepend and may conflict. 
For example, reducing computational complexity does not necessarily lower latency, and minimizing memory usage might not directly reduce energy costs~\cite{liu2023enabling}. 
Accuracy depends on DL model structure and parameters, while latency and energy cost are tied to hardware and system architecture, making optimization not straightforward. 
While existing DL compression and offloading techniques have made progress, they tend to focus on either \textit{algorithm-level} (\eg DL model compression~\cite{bulo2018place} and offloading algorithms~\cite{wang2021context, pang2023adamec}) or \textit{system-level} (\eg computation graphs~\cite{fang2020optimizing}, compilation engines~\cite{lin2020mcunet}, and operator optimization~\cite{niu2021dnnfusion}). 
These methods are not readily extendable to \textit{cross-level dynamic} adaptation. \textit{Handcrafted cross-level optimizations}, such as algorithm-compiler co-designs~\cite{lin2020mcunet}, rely on fixed, \textit{predefined} rules, which are insufficient for handling dynamic mobile environments. 
Therefore, \textit{automated} performance validation and cross-level algorithm-system optimization for model porting are critical yet challenging.

% Despite major advances of \textbf{existing} DL model compression/offloading techniques, they are typically limited to either \textit{algorithm-level}, such as DL model compression~\cite{deng2015reduced, bulo2018place} and offloading algorithms~\cite{wang2021context, pang2023adamec}, or \textit{system-level} scheduling, including computation graphs ~\cite{fang2020optimizing}, compilation engines~\cite{lin2020mcunet}, and operators~\cite{niu2021dnnfusion}. 
% These approaches are not readily extendable to \textit{cross-level dynamic} adaptive optimization. 
% \IEEEpubidadjcol
% \textit{handcrafted cross-level optimizations}, such as algorithm-compiler co-designs~\cite{lin2020mcunet}, rely on fixed, \textit{predefined} rules, which are insufficient for adapting to the dynamic nature of mobile environments.
% Thus, \textit{automated} performance validation and algorithm-system cross-level strategy selection for model porting is essential yet challenging. 

\noindent$\bullet$ \textbf{\textit{Challenge $\#$2}}: 
Model deployment is a non-trivial process even for a single objective.
\textit{Manual} configuration of optimal DL model deployment algorithm and resource scheduling for diverse mobile scenarios places a heavy burden on non-expert developers and lacks the \textit{rapid adaptability}. 
For instance, redesigning DL model architectures for a ReseNet-18 can involve searching from design spaces as large as $2^{102}$, making it impractical for real-time needs.
When multiple model variants are needed, either for different hardware specifications or runtime conditions, manual adaptation becomes labor-intensive and repetitive. 
Previous efforts in model optimization~\cite{liu2020adadeep}, offloading strategies~\cite{lu2019collaborative}, and resource allocation~\cite{wang2022melon} offer solutions, but they often fall short in handling dynamic changes and diverse environments. 
Under resource-limited mobile conditions, such as system overloads or high temperatures triggering DVFS (Dynamic Voltage and Frequency Scaling) to lower processing frequencies, these challenges can degrade DL services and user experience.

\begin{figure*}[t]
\centerline{\includegraphics[width=0.96\textwidth]{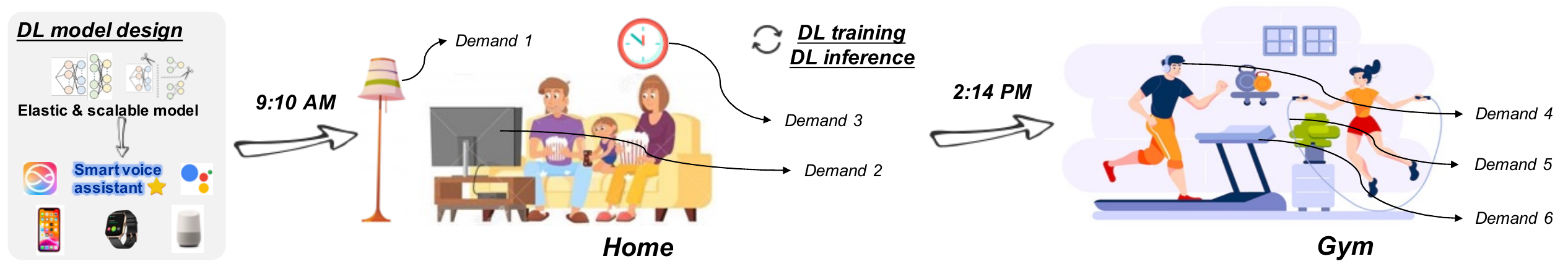}}
\vspace{-1mm}
\caption{Illustration of an example mobile application, \ie an user interacting with a voice assistant model.}
\vspace{-3mm}
\label{fig:example_figure}
\end{figure*}

% 5设计，贡献
Given those challenges and limitations, we present CrowdHMTware, a dynamic context-adaptive DL porting middleware for mobile devices. 
Fundamentally, the core issue lies in the implicit coupling of model design and system deployment.
Mobile developers must balance improving inference accuracy with minimizing memory and computation costs, while mobile users need to maintain high accuracy and accelerate inference. 
Both stages require expertise across DL algorithms and system runtime scheduling, spanning model design, training, and execution phases.
CrowdHMTware decouples these into functional blocks and integrates them into an automated loop, enabling runtime \textit{cross-level} joint optimization.

\noindent \textbf{First}, CrowdHMTware introduces the frontend level optimization through \textit{resource-aware elastic DL compression and offloading}.
We observe that most DL models require retraining when their structure changes, but multi-variant ensemble learning can shift this retraining to the pre-training phase. To leverage this, we propose a runtime adaptive model scaling scheme using a retraining-free network with a \textit{multi-branch backbone} and multiple \textit{compression operators}.
Additionally, we introduce a pre-partitioning strategy based on minimal operator units, decoupling model partitioning from offloading search, ensuring universality across various tasks, models, and devices.
With these inter-/intra-device scaling, the DL models achieve dynamic \textit{scalability} and \textit{divisibility} at runtime, adapting to mobile contexts.

\noindent \textbf{Second}, CrowdHMTware realizes the backend-level optimization in a \textit{dynamic model-adaptive manner} via a compilation engine.
We observe that the accuracy and responsiveness of DL models are bounded by resource availability, even on the same hardware. 
Maximizing hardware utilization without compromising model accuracy can push system performance limits. 
Our \textit{key idea} is that mapping model layers and operators to different memory units in varying sequences impacts latency and resource overhead.
CrowdHMTware addresses this by optimizing at fine-grained levels, \eg computation graph, operator, and memory allocation. 
These strategies also allow more flexible design space for the front-end DL model scaling.

\noindent \textbf{Third}, CrowdHMTware introduces an \textit{automated loop for cross-level co-adaptation} with three extra components, \ie \textit{resource availability monitor}, the \textit{runtime performance profiler}, and the \textit{optimizer}.
We observe that while metrics like computational complexity and memory usage can be derived from the model's dynamic architecture, hardware-dependent metrics such as latency and energy cost are harder to measure due to black-box system architecture and time-variant resource availability. 
To address this, we enhance offline estimation by incorporating resource dynamics.
Also, we employ a lightweight heuristic optimizer, ensuring efficient, adaptive cross-level optimization in dynamic mobile environments.

We implement CrowdHMTware as middleware that leverages existing model specifying and system scheduling strategies, adding system services to improve adaptivity rather than proposing new algorithms.
We evaluate CrowdHMTware across 4 typical mobile applications (including mobile acoustic event awareness, image classification, mobile human activity perception, driver behavior prediction), addressing cross-level system adaptation requests from over 15 mobile and embedded devices. 
Results show that CrowdHMTware achieves the best accuracy-latency tradeoff. 
In dynamic contexts, it reduces latency by up to 10.3$\times$ while improving accuracy by 3.9\%, outperforming state-of-the-art handcrafted/on-demand DL model specification algorithms.
The key contributions of this work are summarized below.
\begin{itemize}
    \item To the best of our knowledge, CrowdHMTware is the first dynamic mobile context-adaptive middleware to realize cross-level co-adaptation, shrinking front-end DL model algorithms and back-end operator/resource scheduling on-the-fly.
    \item CrowdHMTware implements an automated adaptation loop across diverse levels, \ie algorithm level with elastic inference and scalable offloading, and model-adaptive engine level. 
    Also, it involves resource availability monitors, performance predictors, and optimizers for this automation loop.
    \item Using five mobile applications over 15 typical mobile and embedded devices, experiments show that CrowdHMTware can adaptively port diverse DL models to dynamic and diverse mobile deployment environments, outperforming state-of-the-art baselines.
    
\end{itemize}

In the rest of the paper, we present the system overview in \secref{sec:overview}, and elaborate CrowdHMTware's functional block design in \secref{sec:design}. 
We evaluate CrowdHMTware in \secref{sec:evaluation}, review in \secref{sec:related}, and conclude in \secref{sec:conclude}.

\section{Overview}
\label{sec:overview}

\subsection{Problem Study}

There is a growing demand for deep learning (DL)-based mobile systems that are accurate, responsive, energy-efficient, and adaptable to dynamically changing mobile contexts. 
However, previous efforts can not fully meet the dynamic optimization requirements (as discussed in \secref{sec:related}). 
Both the \textit{mobile developer} and the \textit{user} face a significant challenge, how to automatically and efficiently adjust the DL model and its underlying deployment at runtime to meet dynamic and unpredictable demands. 

In particular, the lifecycle of a DL model on mobile devices includes the \textit{design}, \textit{training}, and \textit{execution} phases.
During the \textit{design} phase, developers specify the DL model to fit a target deployment setting. 
However, these settings are often statically modeled, making it difficult to adapt to the runtime dynamics of real-world deployment environments. 
In the \textit{pre-/re-training} phase, developers fine-tune the model to meet new resource constraints, data distributions, and performance goals. 
While crucial for maintaining accuracy, this phase becomes a bottleneck for runtime adaptability.
During the \textit{execution} phase, the pre-trained model interacts with users and the mobile environment, guided by back-end system scheduling strategies. 
This highlights the disconnect between the algorithmic design at the front-end and the system scheduling at the back-end, across different lifecycle stages. 
Offline retraining is often insufficient for real-time adaptability in dynamic mobile systems, highlighting the demand for a holistic, cross-level, and dynamic model deployment strategy.

For instance, \figref{fig:example_figure} shows a scenario in which a user utilizes a smartphone-based voice assistant app (\eg Apple Siri~\cite{Capes2017SiriOD}) that continuously senses voice interaction tasks. 
When application scenarios change in real time (\eg from home to the gym), the voice assistant app experiences different input noisy data, inference requests, and resource statuses.
During usage, the battery-powered mobile device's energy supply is also affected by various factors such as memory access, microphone or camera sensor sampling, and irregular screen activity, further imposing dynamic energy constraints on the deployed DL model. 
Additionally, the memory unit (\eg L2 Cache) is shared by other competing programs, resulting in dynamic availability for DL model parameters.
From the perspective of a mobile \textit{developer}, the DL model might be deployed across different mobile embedded devices, each with varying hardware architectures (computing, memory, and battery resource configurations), DL frameworks (\eg TensorFlow, Pytorch), and constraints similar to those illustrated in the example.

% For example, \TODO{Figure xx} shows an example in which a user carries a  \todo{replace an application like this:  smartphone-based hearing assistant App (\eg xx xx~\cite{}) } to
% sense the \TODO{replace: ambient acoustic event of interest} continuously. 
% %
% During its use, the device’s battery is dynamically consumed during the long-term execution, the memory access, the microphone/camera sensor sampling, and the screen with unpredictable frequency, which further characterize the dynamic energy constraints for the deployed DL model. 
% %
% And the storage unit (\eg L2-Cache) is also dynamically occupied by other applications, resulting in various memory budgets for DL model parameters. 
% Also, 

To sum up, a dynamic context-adaptive DL model deployment middleware for heterogeneous mobile devices with \textit{diversity} and \textit{dynamic} nature must possess the following capabilities:
\begin{itemize}
    \item \textbf{Compatibility with Heterogeneous Execution Environment}: Mobile and embedded devices vary widely, from embedded sensors and IoT devices to smartphones and wearables, each with different hardware capabilities. 
    Even for the same DL model inference task using the same framework, latency can differ significantly, \eg inference time on Raspberry Pi is $3\times{}$ that of Jetson Nano. 
    When running MobileNet inference, Raspberry Pi 4 may take 615 ms, while Jetson Nano only takes about 202 ms. 
    CrowdHMTware supports a variety of \textit{heterogeneous DL frameworks}, such as PyTorch~\cite{paszke2019pytorch} and MACE~\cite{jia2022codl}, and runs on \textit{heterogeneous processors}, including CPUs, GPUs, and NPUs. 
    It addresses performance disparities and \textit{operator compatibility} challenges arising from diverse hardware, software, DL models, tasks, and performance requirements.
    \item \textbf{Fast Response to Runtime Dynamics}: Runtime hardware resource availability on mobile devices is naturally impacted by various runtime factors like \textit{processor temperature}, dynamic voltage and frequency scaling (\textit{DVFS}), competing processes, and \textit{processor utilization}. 
    \textit{Adjustments} by any \sysname component also influence these dynamics. 
    Thus, adapting to \textit{dynamic demands on performance and resource efficiency} is desired, such as changing energy budgets as battery levels decrease.
\end{itemize}
CrowdHMTware enhances runtime front-/back-end cross-level optimization by automating the tuning of model structures, offloading strategies, operator execution, and memory scheduling, adapting to user demands and hardware capabilities.

\begin{figure*}[t]
  \centering
  \includegraphics[width=0.9\linewidth]{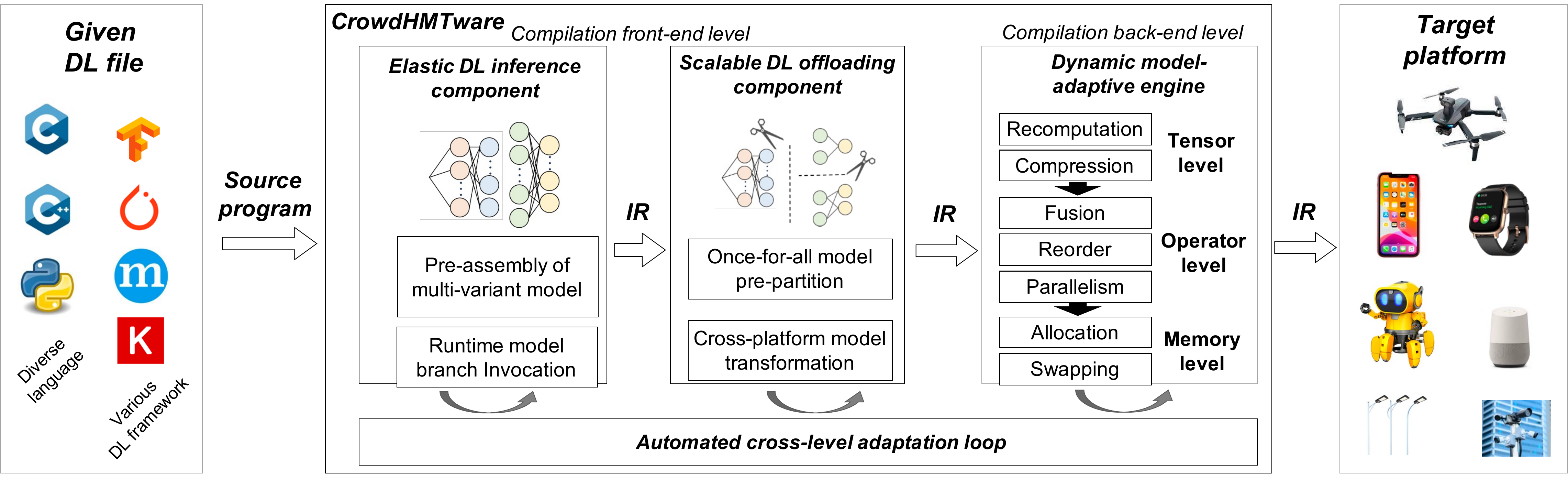}
  % \vspace{-2mm}
  \caption{Illustration of the CrowdHMTware architecture.}
  \vspace{-2mm}
  \label{fig:overview}
\end{figure*}

\subsection{System Overview}
To address these challenges, CrowdHMTware comprises three key components as depicted in \figref{fig:overview}: the \textit{elastic DL inference} component, the \textit{scalable DL offloading} component, and the \textit{model-adaptive back-end compilation engine}.
Specifically, the \textit{elastic DL inference component} enhances runtime flexibility, adapting models to dynamic conditions without retraining, and avoiding accuracy degradation during adaptations (see \secref{sec:elastic_model}).
The \textit{scalable DL offloading component} enables on-demand distributed offloading to alleviate local resource constraints and improve inference efficiency. 
It decouples hierarchical pre-partitioning and performance-aware offloading strategies, integrating operator optimization into the cross-framework conversion process (see \secref{subsec:offloading}).
The \textit{model-adaptive back-end engine} optimizes resource scheduling, minimizes fragmentation, and enhances data reuse by tailoring operator and resource scheduling to dynamic model structures (see \secref{sec_engine}).
CrowdHMTware also features an \textit{automated adaptation loop} that integrates these components effectively. 
This loop includes a resource availability monitor for continuous hardware tracking, a runtime performance profiler that factors resource dynamics into energy and latency estimations, and an optimizer that dynamically adjusts above operations to meet the evolving needs of mobile applications efficiently (see \secref{subsec:loop}).

CrowdHMTware implements this \textit{runtime cross-level optimization} as a middleware, optimizing operational conditions and improving application delivery and user satisfaction across various deployment scenarios.
Specifically, it enhances dynamic DL-based application performance management, facilitating robust DL deployment (we defer more details in \secref{subsec:implementation}).
\section{CrowdHMTware Design}
\label{sec:design}

\subsection{Front-end level: Resource-aware Elastic DL Inference}
\label{sec:elastic_model}

\textbf{\textit{Limitations of Existing Methods}}.
Despite advancements in DL model compression algorithms~\cite{liu2020adadeep}, existing methods lack an automated integration with the \textit{runtime flexibility} needed to adapt to the context dynamics of continuously running mobile applications.
A key challenge for both mobile developers and users is determining \textit{how to dynamically adapt DL model compression at runtime}. 
Traditional compression techniques typically do not support re-expanding a compressed model easily.
In response, we introduce a pre-trained, multi-variant DL model, which is based on a backbone network equipped with multiple compression operator variants, allowing dynamic adjustment of compression levels in response to changing runtime conditions.

\subsubsection{Pre-assembly of Multi-variant Operators}

CrowdHMTware employs a multi-branch backbone with multi-variant compression operators to enhance \textit{adaptability}. 
In particular, the \textit{backbone} architecture incorporates an \textit{early-exit multi-branch} model optimized for efficiency, beginning with convolutional layers that downsample features to reduce spatial dimensions and data volume while retaining output channels to preserve essential feature information. 
\rev{Each branch is equipped with an \textit{adaptive early-exit mechanism}, where the decision to exit is based on confidence thresholds derived from intermediate feature representations.}
% These thresholds are dynamically adjusted to align with task-specific accuracy requirements and available device resources.
\textit{Adaptive average pooling} layer then resizes the feature map to reduce parameter size. 
\textit{Dropout} layers prevent overfitting and improve generalization. 
Additionally, downsampling in the early-exit branches significantly lightens the data load on the densely parameterized fully connected layer.

Additionally, we incorporate six categories of typical compression \textit{operators}, \rev{\ie $\eta_1\sim \eta_6$}, upon the backbone network that integrates multiple scaling dimensions, such as width, depth, and connection. 
Specifically, it includes a set of coarse-grained compression operators (\eg Fire~\cite{iandola2016squeezenet}, SVD-based~\cite{wu2018deep}, sparse coding-based~\cite{bhattacharya2016sparsification} factorization), \textit{for faster convergence}, and the fine-grained compression operators (\eg channel-level and depth-level pruning ~\cite{cai2019once} and channel-wise randomization ~\cite{yang2022progressive}), \textit{for better diversity}.
This allows for flexible tailoring of the model to meet specific dynamic performance requirements, thereby maximizing efficiency across diverse conditions.

\begin{itemize}
\item $\eta_1$: \textit{low-rank convolution factorization} operators (\eg SVD-based~\cite{wu2018deep}), sparse coding-based ~\cite{bhattacharya2016sparsification} factorization, or depth/ group-wise convolution~\cite{li2019dabnet} decompose a conv layer into several layers with smaller kernel size. 
\item $\eta_2$: \textit{multi-branch channel merging} operators (\eg Fire ~\cite{iandola2016squeezenet}) increase the model depth with fewer parameters by replacing a conv layer using squeeze and expand layer.
\item $\eta_3$: \textit{composite convolution scaling} operators (\eg in EfficientNet~\cite{koonce2021efficientnet}) adjust kernel size, stride, and the number of input and output channels. 
\item $\eta_4$: generates \textit{basic feature maps} (\ie Ghost module) with a few convolution operations and then expands them using linear operations(\eg in GhostNet~\cite{han2020ghostnet}). 
\item $\eta_5$: \textit{depth-wise scaling} operators (\eg depth-elastic pruning ~\cite{cai2019once}, residual connection ~\cite{he2016deep}) derive a shallower variant-DL model from a backbone-DL model via skipping connections.
\item $\eta_6$: \textit{channel-wise scaling} operators (\eg channel-level pruning ~\cite{cai2019once} and channel-wise architecture noise injection ~\cite{yang2022progressive}) can tune variable operator-variant sampling.
\end{itemize}
% \lsc{These multi-variant operators are cost-effective and well-suited for dynamic scenarios requiring quick responses.}

\textit{\textbf{Ensemble Learning of Backbone and Multi-variant Operators (offline)}}.
\rev{To address the challenge of performance degradation and uncertainty in guaranteeing diversity when only partial variants are selected during inference due to limited device resources, we propose enhancements to the dynamic inference process. Specifically, we eliminate the need for weight retraining by moving this process ahead into the ensemble training phase of the multi-variant model. 
The multi-variant model training integrates a backbone network with multiple variant networks derived through various compression operators. 
Additionally, we introduce \textit{weight recycling} across diverse variants to prevent catastrophic interference or degradation when only partial operators are selected during inference~\cite{cai2018efficient}. }
The process begins with a high-performance, intricately designed deep learning (DL) model, including both its architecture and weights. 
We then adjust the model width and connection paths to develop effective new variants. 
\rev{However, ensemble training presents challenges, as adjustments may cause variants to deviate from the original backbone configuration, potentially \textit{disrupting or overwriting} the weights of other variants, which can compromise overall performance.}
To mitigate these issues, we first ensure that the backbone DL model achieves high accuracy using standard back-propagation. 
Subsequently, we refine the training of compression operators through parameter transformation techniques~\cite{zhang2019parameter} (\eg $\eta_1$, $\eta_2$), knowledge distillation~\cite{gou2021knowledge} (\eg $\eta_4$, $\eta_5$), and channel-wise mutation techniques~\cite{cui2019chip} (\eg $\eta_6$). 
These approaches collectively enhance model adaptability and maintain performance consistency across all variants.

% \rev{To enhance the dynamic inference process and eliminate the need for retraining weights when only selecting partial variants, we put the retraining process ahead in the ensemble training of the multi-variant model to get rid of weight retraining during dynamic inference. 
% %
% Specifically, the multi-variant model training is an ensemble of a backbone-net and multiple variant-nets derived by various compression operators.}
% %
% And we leverage \textit{weight recycling} across diverse variants to prevent catastrophic interference or only partial operators are selected during inference time~\cite{cai2018efficient}. 
% %
% We start with a high-performance, intricately designed DL model (encompassing both architecture and weights), and then adjust model width and connection paths to develop an effective new model variant.
% However, this approach introduces challenges during ensemble training. 
% Adjustments may cause variants to stray from the initial backbone configuration, potentially \textit{disrupting or overwriting} weights of other variants, thereby impacting overall performance.
% To mitigate this, we first ensure the backbone DL model achieves high accuracy through standard back-propagation. 
% Subsequent steps involve employing parameter transformation techniques~\cite{zhang2019parameter} for $\eta_1$, for $\eta_2$ and knowledge distillation ~\cite{gou2021knowledge} for $\eta_4$, for $\eta_5$ to refine learning compression operators, along with channel-wise mutation techniques~\cite{cui2019chip} for $\eta_6$ to further enhance model adaptability.

\subsubsection{Runtime Parameter Adaptation for Handling Data Drift}
We defer the runtime selection and combination of these variant operators in the context-aware automated adaptation loop (see \secref{subsec:loop}).
\rev{While we avoid retraining due to model structural compression changes, CrowdHMTware also incorporates test-time adaptation to address data shift issues. 
The shift between the distribution of testing data and training data can lead to a decline in inference accuracy.
This unsupervised adaptation strategy dynamically adjusts pre-trained parameter weights to unlabeled live data during testing, enhancing the accuracy for target data without needing access to source data or supervision from the pre-training phase. 
Integrating test-time adaptation for selective weight updating is particularly valuable in mobile applications, where accessing source datasets may be impractical due to privacy concerns or bandwidth limitations, and annotating target domain data can be both challenging and labor-intensive.
}

\subsection{Front-end level: Resource-aware Scalable DL Offloading}
\label{subsec:offloading}

Since the accuracy of the above-mentioned on-device DL inference is still \textit{bounded} by resource availability, supplemental \textit{on-demand scaling} to the distributed scheme can be utilized. 
Specifically, deploying different decoupled parts of DL models across multiple devices can reduce local resource demands and boost inference efficiency. 
This is especially advantageous given the complex structure of modern DL models, \eg ResNet-152 with 228MB parameters~\cite{zhang2022novel}, and DistilBERT with 251MB parameters~\cite{sanh2019distilbert},  which demand significant memory and computational resources.

\begin{figure}[t]
\centerline{\includegraphics[width=0.5\textwidth]{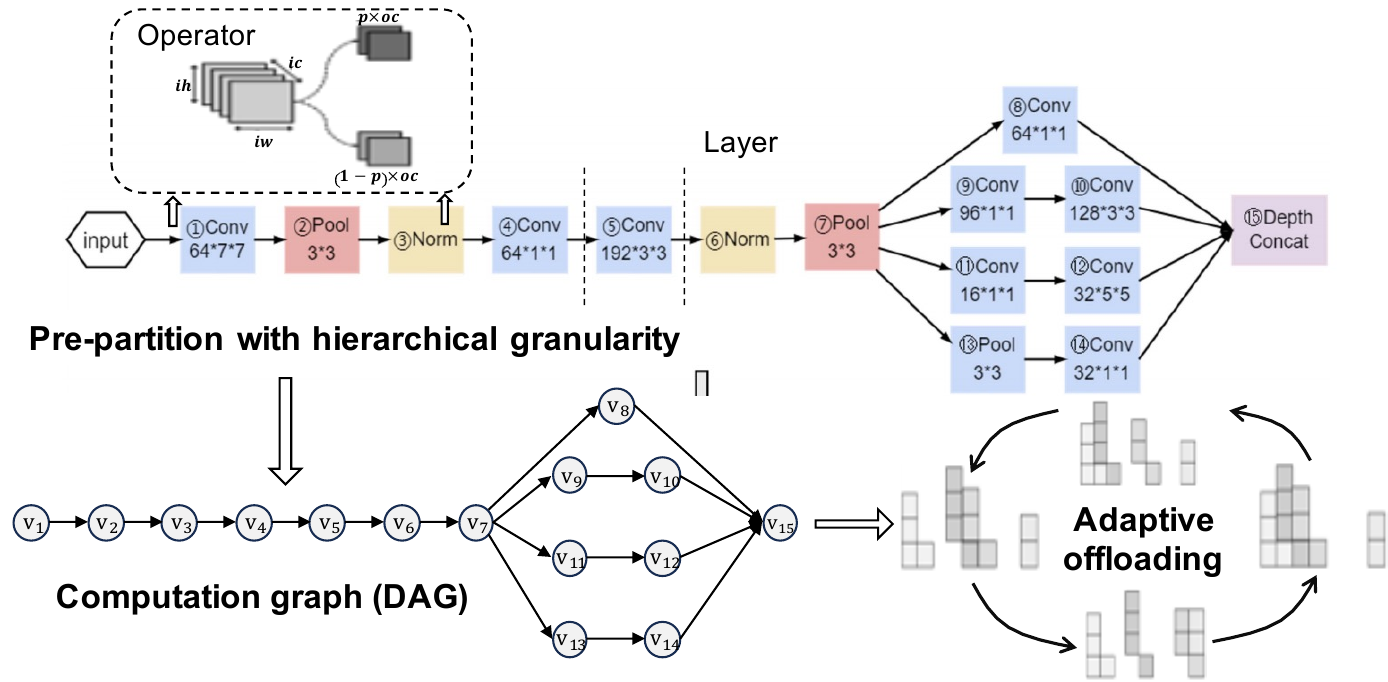}}
\vspace{-1mm}
\caption{Model pre-partitioning with hierarchical granularity and adaptive offloading through pre-partition combination.}
\vspace{-3mm}
\label{fig_cal_cal}
\end{figure}

\subsubsection{Operator-based DL model Pre-partition}
To facilitate dynamically adaptive DL model inference, CrowdHMTware middleware leverages a \textit{pre-partition strategy} that operates independently of specific application latency requirements and device resource constraints. 
This is based on hierarchical hybrid granularity that combines both computational graph and operator levels, ensuring universality \textit{across different tasks, DL models,
and DL frameworks} (\eg Tensorflow Lite~\cite{tensorflowlite}, Pytorch~\cite{imambi2021pytorch}, MCNN~\cite{tripathi2020mcnn}).
In particular, our approach is driven by two key principles:
\begin{itemize}
    \item \textit{i) Uniform operator range}: by focusing on the stable operational ranges of key operators like convolution, pooling, and fully connected layers, we address the inherent heterogeneity and dynamism of DL models, ensuring consistent performance.
    \item \textit{ii) Granular computational graphs}: higher-level granularity within the computational graph simplifies the creation of a compact and adaptable search space. 
    This reduces the complexity of locating optimal partition points for operators, facilitating swift adjustments to the offloading strategies in any mobile scenario and promoting seamless integration with elastic scaling techniques.
\end{itemize}

\textbf{\textit{Hierarchical Decoupling of Operators}}.
To enhance scalability and maximize operator parallelism in the DL model pre-partition phase, CrowdHMTware employs a \textit{hierarchical decoupling} approach, as shown in \figref{fig_cal_cal}. 
\rev{The model is first segmented at the \textit{operator} level for flexibility, followed by topological sorting to create independent \textit{operation flows}. 
These flows link operators to their input/output tensors using sparse matrix mappings, minimizing idle time and enabling parallel execution. 
% This hierarchical granularity optimizes memory and performance, balancing resource allocation and reducing latency for efficient operation in dynamic, resource-constrained environments.
}
% These partitions are organized into operation flows within a Directed Acyclic Graph (DAG).
% Initially, it segments the DL model at the primitive \textit{operator} level for flexibility and adaptability. 
% And then, using topological sorting, the graph is divided into multiple independent \textit{operation flows}, each represented as sparse matrix mapping operations to their respective input or output tensors. 
% This scheme can minimize idle time, enabling continuous operation and parallel processing to boost performance in dynamic environments.
%

\textit{\textbf{Adaptive Cross-device Operator Offloading.}}
For dynamically optimal task offloading, CrowdHMTware employs a graph-based search algorithm to efficiently determine the optimal combination of pre-partitioned DL model components for deployment across various devices. 
It selects the most effective sequence for combining these components, optimizing latency, and quickly adapting to changes in deployment contexts such as resource availability and latency requirements.

\begin{figure}[t]
\centerline{\includegraphics[width=0.5\textwidth]{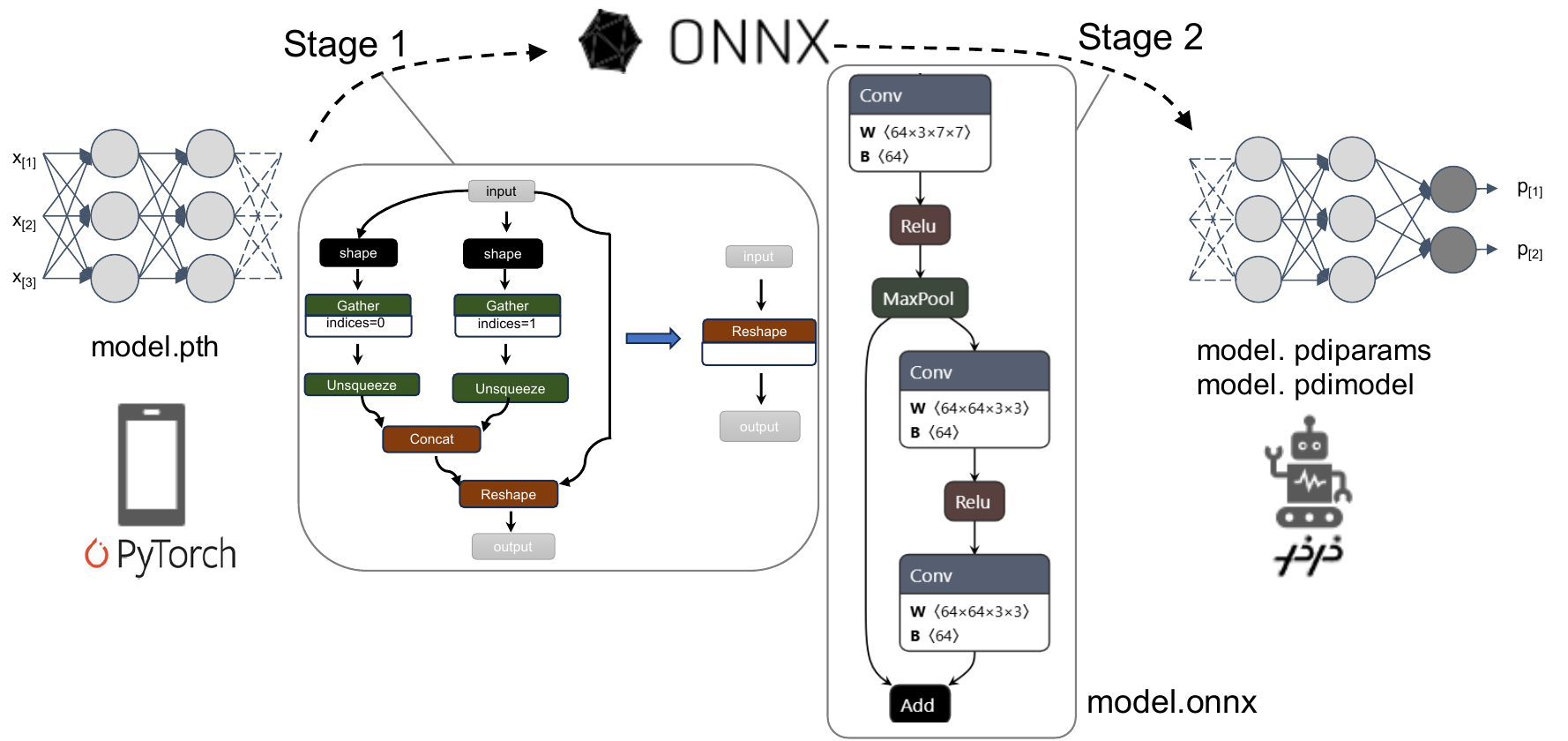}}
% \vspace{-2mm}
\caption{Integrating operator optimization into cross-framework transformation process, \eg from PyTorch to PP.}
\vspace{-4mm}
\label{fig_operator_onnx}
\end{figure}

\subsubsection{Redundance-aware Cross-platform DL Model Transformation}
Given the diversity of popular DL frameworks like PyTorch, TensorFlow, and Paddle, a significant challenge exists in real-world offloading scenarios due to the different frameworks employed by mobile and embedded devices. 
Typically, these devices cannot directly execute a model file offloaded from one framework (\eg TensorFlow file at device A) to another (\eg Pytorch execution at device B). 
To overcome these interoperability challenges, the ONNX tool offers a standardized approach for storing and transferring DL models across various systems, facilitating model execution across different frameworks~\cite{lin2019onnc}. 

However, the ONNX does not automatically address the \textit{redundancy of operators} that can \textit{occur} during the transformation process. 
Framework heterogeneity can lead to redundant operators in the same DL model post-compilation.
In view of this challenge, we propose a two-stage conversion process within the ONNX tool that includes operator optimization, as illustrated in \figref{fig_operator_onnx}. 
\rev{The DL model is first converted into an intermediary computational graph format, where operator dependencies and data flow directions are analyzed. This analysis identifies opportunities for operator fusion (\eg merging BatchNorm with convolution layers) and removes duplicate or redundant operators while ensuring the computation remains unchanged. This process guarantees that critical computational steps contributing to the model’s accuracy are preserved.
In the second stage, 
the optimized graph undergoes a global traversal to assess each node's dependency on its inputs. Operators are classified as either dynamic (dependent on runtime inputs) or constant (whose outputs remain static regardless of inputs). Constant operators are further analyzed to determine their role in the overall computation. If deemed redundant, these operators are removed or replaced by precomputed values, ensuring that simplifications do not affect model accuracy.
}
% a global traversal of the data flow assesses each node’s dependency on its inputs to determine dynamic and constant operators. 
% Constant operators with outputs that do not vary with inputs are identified as redundant and replaced, simplifying the ONNX graph by removing these nodes.

\subsection{Back-end level: Model-adaptive Compilation Engine}
\label{sec_engine}
Given the dynamic nature of the DL model structure derived by the upper-level middleware component (as mentioned in $\S$ \ref{sec:elastic_model}), CrowdHMTware employs a dynamic model-adaptive engine to optimize resource scheduling, minimizes resource fragmentation, and enhances data reuse, as shown in \figref{fig:engine-overview}. 
Specifically, CrowdHMTware's engine focuses on three key aspects during runtime backend compilation:
\textit{i) Computation graph optimization} involves reordering or fusing operators within the computation graph to reduce memory usage and decrease memory I/O access delays in a model-aware manner.
\textit{ii) Hardware resource scheduling} targets optimal utilization of hardware resources at a fundamental level close to the hardware.
\textit{iii) Intermediate activation management} during weight adaptation helps reduce peak memory demands by considering the tensor lifecycle.
We elaborate on their design next.

\begin{figure}[t]
  \centering
  \includegraphics[width=0.92\linewidth]{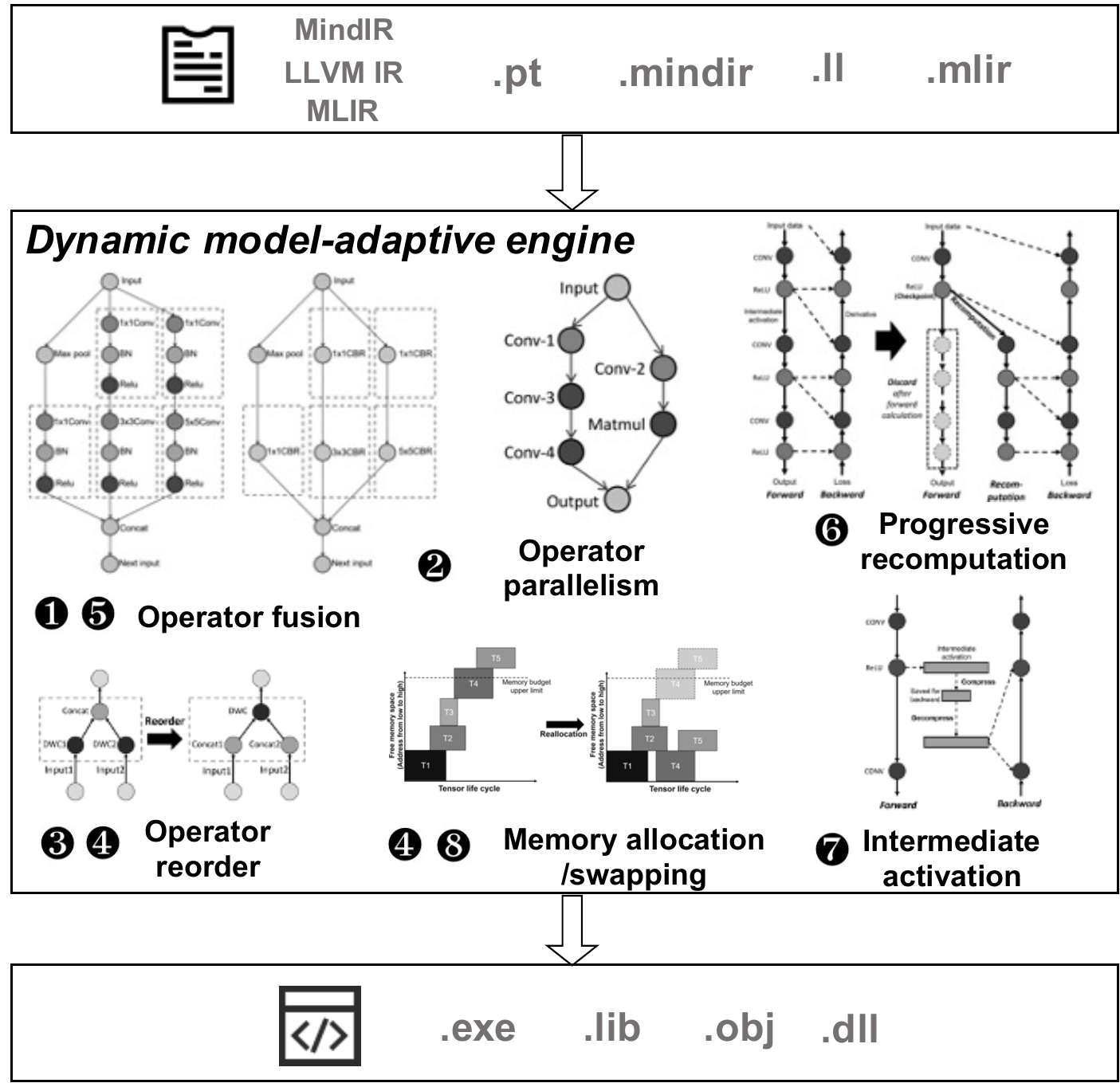}
  % \vspace{-2mm}
  \caption{Illustration of the dynamic model-adaptive engine.}
  \vspace{-2mm}
  \label{fig:engine-overview}
\end{figure}

\subsubsection{Compilation Engine for Elastic Inference.}
The above model specification component struggles to achieve near-/realtime inference on resource-scarce devices. 
This difficulty stems from the limitations imposed by the runtime availability of resources. 
To address this, CrowdHMTware includes a dynamic model-adaptive engine to improve resource availability.

\textbf{\textit{\ding{182} Runtime Operator fusion.}}
Intermediate feature maps, as inputs for multiple operators in computation graphs, can significantly increase memory usage and processing delays. 
To tackle this, fusing adjacent operators into a single one is effective~\cite{niu2021dnnfusion}. 
While DL frameworks like TensorFlow Lite, Mace, Pytorch-Mobile offer APIs, CrowdHMTware goes further by integrating five operator fusion strategies, extending fixed fusion patterns to dynamic settings, which enhance runtime efficiency.
\rev{Specifically, five operator fusion strategies include linear fusion, convolution-batchNorm fusion, element-wise operator fusion (\eg for ReLU, Sigmoid, or Tanh), channel-wise fusion (\eg pointwise convolutions), and reduction fusion (\eg sum, mean, or max pooling).
}
Specifically, it classifies the operators based on the mapping relationships between their inputs and outputs, and during runtime, it progressively attempts operator fusion across different types to increase fusion opportunities.
This allows CrowdHMTware to reduce the memory footprint of intermediate feature maps and decrease delays by minimizing the number of separate operations that need to be managed and executed.

\textbf{\textit{\ding{183} Cross-core Operator parallelism.}}
Modern mobile and embedded platforms often feature heterogeneous processors, \eg multi-core CPUs, GPUs, and NPUs, which show speed variances and require customized optimization strategies to speedup and address memory access bottlenecks. 
% For instance, CrowdHMTware involves CPU cache mechanisms to reduce memory access delays, easing bandwidth constraints. 
% On GPUs, it exploits thread parallelism, allowing multiple threads to operate concurrently. 
% This ensures that even if some threads stall, others continue, keeping the GPU active. 
CrowdHMTware implements inter-operator parallelism to allow simultaneous execution of multiple operators, significantly boosting system efficiency and responsiveness. 
Specifically, it achieves efficient data sharing between the CPU and GPU, and by integrating a lightweight and precise latency prediction model (we defer more details in \secref{subsec:loop}), it enables cross-core operator parallelism, thereby improving concurrent execution and optimizing resource utilization.

\textbf{\textit{\ding{184} Tensor Lifetime-aware Memory allocation.}}
CrowdHMTware optimizes memory allocation, crucial for DL tasks closely tied to hardware resources. 
% It manages a tensor's lifecycle from creation to final use to maximize device memory utilization, arranging these lifecycles in the memory graph without overlap . 
\rev{Specifically, as shown in \figref{fig:engine-overview}, it analyzes the lifecycle of each tensor in the computation graph, from creation to final use, arranging these lifecycles in the memory graph to avoid overlapping allocations. 
Next, it establishes global lifecycle constraints, considering data flows and operator dependencies in the computation graph to optimize the order of tensor allocation. 
Finally, heuristic algorithms are applied to resolve conflicts and enable memory reuse, dynamically adjusting the allocation plan by prioritizing the reuse of idle memory blocks. 
% This approach minimizes memory fragmentation and significantly improves memory utilization under static hardware constraints.
}

% Unlike traditional DL frameworks that use locally optimal greedy algorithms like BestFit~\cite{abadi2016tensorflow}, CrowdHMTware integrates computation graph flows, data, and operator dependencies. 
% It sets global tensor lifecycle constraints and uses heuristic algorithms to create an adaptive memory plan that significantly enhances memory efficiency within static limits.

% CrowdHMTware improves memory allocation, a critical level closely integrated with hardware resources for DL tasks. 
% It strategically manages a tensor's "life cycle"—from creation to final use—to determine its optimal occupancy in device memory, aiming to maximize space utilization. 
% These life cycles, depicted as rectangular blocks in the memory lifecycle graph (refer to \figref{fig:overview}), are organized without overlap.
% Traditional DL frameworks~\cite{abadi2016tensorflow} often rely on greedy algorithms like BestFit, which, while effective at providing locally optimal solutions, fail to achieve runtime optimization. 
% Addressing this shortfall, CrowdHMTware analyzes and integrates the parallel flows of the computation graph, along with data and operator dependencies. 
% By establishing global tensor lifecycle constraints and employing heuristic algorithms, CrowdHMTware derives an adaptive memory plan to enhance memory efficiency towards static limits. 

\subsubsection{Compilation Engine for Test-time Weight Adaptation.}
\lsc{
DL test-time adaptation is more complex than inference due to the need to retain intermediate activations until the completion of gradient computations in the backward pass, leading to substantial memory use and idle time. 
To this end, CrowdHMTware optimizes operator execution, minimizes idle time, and reallocates resources.
}

\textbf{\textit{ \ding{185} Operator reordering during backpropagation.}}
In traditional DL training, gradients are retained throughout backpropagation for uniform weight updates after all gradients are computed. 
CrowdHMTware modifies this by reordering operator execution in the computation graph during backpropagation, swapping the sequences of gradient computation and weight updating. 
\rev{To address the dependency issue where the gradient of one operator relies on another, CrowdHMTware enforces dependency-aware constraints in the reordering process.}
This allows for the immediate disposal of each gradient right after its corresponding layer update.
% , eliminating the need to retain it throughout the training iteration.

\textbf{\textit{\ding{186} Operator fusion during backpropagation.}}
\lsc{By fusing two adjacent operators in the computational graph, the intermediate activation from the first layer can be directly utilized by the next, eliminating unnecessary memory transactions during propagation. 
CrowdHMTware incorporates a precise memory cost model to assess and optimize potential operator fusion schemes, adaptable to online deployment scenarios. 
}

\textbf{\textit{\ding{187} Progressive recomputation}. }
CrowdHMTware optimizes memory usage in dynamic memory pools by introducing a progressive recomputation technique. 
It adjusts tensor positions to minimize interference from discarded tensors and proactively discards tensors when memory exceeds thresholds. 
When a new memory budget is detected, CrowdHMTware rapidly adjusts by recomputing and retaining necessary tensors, realigning their locations to align with the overarching adaptation strategy before training.
\begin{figure*}[t]
\centerline{\includegraphics[width=0.8\textwidth]{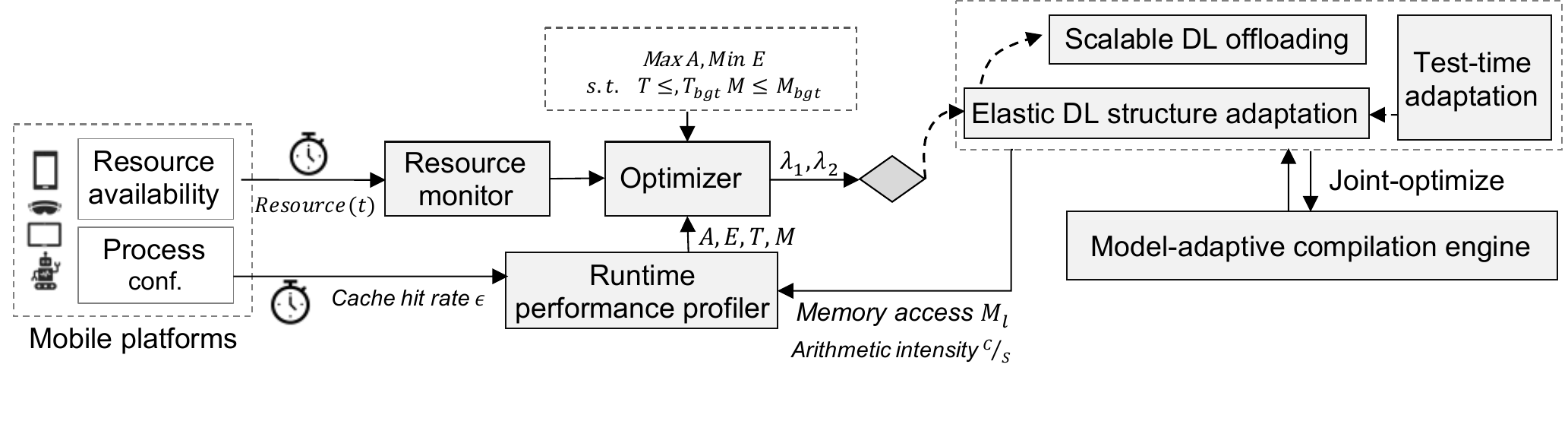}}
\vspace{-2mm}
\caption{Illustration of CrowdHMTware's automated loop for cross-level adaptation.}
\vspace{-2mm}
\label{fig_over_workflow}
\end{figure*}

\textbf{\textit{\ding{188} Intermediate activation compression.}}
CrowdHMTware compresses intermediate activations post-forward pass and decodes them for backpropagation, optimizing both latency and memory usage. It quickly releases feature maps during the forward pass to conserve memory and employs layer-specific lossless and lossy encoding strategies. By storing feature maps from pooling to ReLU layers in 4/8-bit rather than 32-bit formats, CrowdHMTware substantially reduces memory requirements.

\textbf{\textit{\ding{189} Model-adaptive Memory Swapping.}}
\lsc{
CrowdHMTware optimizes memory resource by managing intermediate activations only necessary during forward and backward passes, efficiently swapping these between fast (\eg GPU) and slower, larger memory (\eg CPU) as needed. This strategy not only speeds up computations but also lowers memory demands. 
Unlike typical operating system memory swapping, which can introduce delays when using external memory, we find that DL inference’s sequential computation allows for tailored swapping strategies. 
CrowdHMTware dynamically selects convolution operators, adjusts mini-batch sizes, and adjusts memory-swapping tactics to improve performance, employing advanced techniques like fast Fourier transforms and Winograd algorithms for further efficiency. 
It also mitigates potential accuracy losses from reduced batch sizes by dividing batches into smaller sub-batches according to available memory, accumulating gradients to maintain accuracy.
}

% \subsection{Putting Them Together}

\subsection{Automated Loop for Cross-level Co-adaptation}
\label{subsec:loop}
\figref{fig_over_workflow} outlines the workflow of the dynamic context-aware adaptation controller integrated within the middleware. 
This system operates through an \textit{automated} adaptation loop comprising three \textit{extra} core components: the \textit{resource availability monitor}, the \textit{runtime performance profiler}, and the \textit{optimizer}.
The \textit{resource monitor} keeps track of computing and memory resources available within and across devices. 
Simultaneously, the \textit{runtime performance profiler} assesses DL model-related and hardware-dependent performance metrics under the current system configurations. 
CrowdHMTware initially prioritizes an on-device inference scheme but will consider offloading tasks to other devices if resource constraints and performance cannot be met on-device.
Specifically, if resource demands exceed what the device can supply, the analyzer informs the optimizer.
It then adjusts by selecting alternative DL compression variants (\secref{sec:elastic_model}), offloading (\secref{subsec:offloading}), and reshaping operator and resource scheduling (\secref{sec_engine}).
This automated control loop continuously monitors system changes and executes adaptations at a predefined frequency (\eg per second), ensuring the middleware dynamically aligns with the current context for optimal performance.

\subsubsection{Runtime Performance Profiler}
Precise and timely estimation of performance metrics is crucial for scaling the DL structure hyperparameter and deployment strategy at runtime. 
% Performance metrics for a DL model include both model-related metrics, such as computational complexity $C$ and memory usage $M$, and hardware-dependent metrics like latency $T$ and energy cost $E$. 
While model-related metrics (\eg computational complexity $C$ and memory usage $M$) can be calculated from the dynamic architecture of the model, hardware-dependent metrics (\eg latency $T$ and energy cost $E$) are challenging to measure due to their reliance on system architecture and fluctuating resource availability.
Traditionally, estimations of hardware-dependent metrics like energy cost and latency are conducted offline, using tools such as NN-meter~\cite{zhang2021nn} and CoDL~\cite{jia2022codl}, which generally presume static or ample resource scenarios. 
This approach often leads to inaccuracies under dynamically changing resource conditions. 
Factors such as CPU/GPU temperature, the number of competing processes, OS scheduling policies, and adjustments made by CrowdHMTware middleware, including elastic inference and resource scheduling, significantly impact resource availability. 
For example, mobile CPU and GPU OS kernels typically employ round-robin scheduling for multiple processes, and Dynamic Voltage Frequency Scaling (DVFS) is often activated to reduce clock frequency and prevent overheating.
To bridge this gap, we enhance offline estimation models by incorporating the dynamics of resource availability.

\textit{\textbf{Runtime Energy Cost $E$ Estimation}}.
Modeling the energy cost of a DL model on dynamically changing hardware is inherently complex. To more effectively estimate energy consumption, especially on resource-limited mobile devices that process networks layer by layer, we break down the energy cost into individual layers. We simplify this process by focusing on the cache-hit-rate \textit{$\epsilon$}, a parameter that can be directly measured during runtime. 
This is grounded in the fact that energy costs are predominantly driven by DRAM accesses, which mainly occur during cache misses. 
By utilizing the cache-hit-rate $\epsilon$ as a multiplier, we can estimate the energy cost as follows:
\begin{equation} \label{eq:E_l}
E = \sum_{l=1}^N \delta_1 \times C_l + \epsilon \times \delta_{2} \times M_l + (1 - \epsilon) \times \delta_{3} \times M_l + M_l \times \delta_{SM},
\end{equation}
where computation cost $C_l$ of layer $l$ can be formulated by the number of MACs.
% \(M\) represents SRAM access cost.
Although it is challenging to measure $\delta_1, \delta_2$, and $\delta_3$ separately, their ratio can be measured offline for specific platforms.
$N$ is the layer number, which is dynamically determined during elastic inference.
$\sigma_1$, $\sigma_2$, $\sigma_3$, $\sigma_{SM}$ are the unit energy cost of MAC computation, cache access, DRAM access, and shared memory access. 
These coefficients can be measured offline. 
We empirically set $\sigma_1$:$\sigma_2$:$\sigma_3$:$\sigma_{SM}$ = 1:6:200:2 for mobile GPU platforms. 
As for mobile CPU platforms , $\sigma_{SM}$ since they do not have such shared memory space, and thus $\sigma_1:\sigma_2:\sigma_3$ =1:6:200. 
The coefficient ratios are stable for a specific hardware architecture. 
According to our evaluations, these parameters work with different DNN inference frameworks, \textit{e.g.,} Raspberry Pi 4B (CPU) + NCNN, Hornor 9 (CPU) + Pytorch Mobile, and Nvidia Jetson Nano (GPU). 
Additionally, this model unit-based approach is applicable to both CNN and transformer models, they have different model units. 
For example, the basic units in ResNet include Conv2d, BatchNorm, and Linear. 
In a Transformer model, the basic units consist of projectors Q, K, V, LayerNorm, and the feed-forward network (FFN).

\textit{\textbf{Runtime Latency $T$ Estimation}:} 
The inference latency of DL models on mobile devices is heavily influenced by the system (\eg operator) scheduling and device memory hierarchy. 
Current latency profiling methods, such as \eg linear regression based on the number of computations (\ie FLOPs) \cite{liu2018darts}, complex black-box machine learning \cite{zhang2021nn}, and platform-aware methods (\ie concurrency) \cite{jia2022codl}, are developed offline and assume static conditions. 
However, the dynamic and variable mobile environment significantly impacts the accuracy of these models, highlighting a need for profiling methods that adapt in real-time to non-stationary operational conditions.
We discretize latency at the layer level, where the execution latency of a DL model mainly comes from computation ($C$) and memory access ($M$). We relate the memory access coefficient to the cache-hit-rate ($\epsilon$), similar to the energy metric, because sharing limited cache among processes increases memory access latency. %
Additionally, in the case of a cache miss, there is extra latency for data movement, determined by the bus bandwidth ratio.
We use dynamic arithmetic intensity $\delta$ as a proxy for the degree of reuse of parameters and activations and the time required for processing inputs. 
The arithmetic intensity $\delta$ of a dynamically scaled DL model can predict how efficiently arithmetic operations reuse data fetched from different levels in the memory hierarchy and how efficiently the arithmetic operation is executed. 
Since $\delta$ represents the ratio of computation to memory cost ($C/M$), we integrate it into the $\lambda_1$ coefficient to represent the $\lambda_1/\lambda_2$ ratio.
Thus, the latency is computed by:
\begin{equation} \label{eq:T_l}
T = \sum_{l=1}^N  \lambda_1 \ \times \delta_l \times C_l + \epsilon \times \lambda_{2} \times M_l + (1 - \epsilon) \times \lambda_{3} \times M_l
\end{equation}

In particular, CrowdHMTware's profiler works in offline and online stages.
\begin{itemize}
    \item \textit{Offline Stage}.
    The unit energy costs $\delta_1 \sim \delta_3$, and the unite latency $\lambda_1 \sim \lambda_3$, and the MAC throughput at 100\% cache-hit-rate, are measured offline for the given platform.
    During the offline data analysis stage, we use a digital power monitor (\eg Monsoon AAA10F) to sample the actual power cost by profiling the device through its external power input. 
    The energy cost of accessing the cache, DRAM, and shared memory, normalized to that of a MAC operation, is determined based on these. 
    The MAC execution rate per second at 100\% cache-hit-rate is the hardware-specified computation frequency times the parallelism (\eg 16 bits). 
    For latency record collecting, we embed a timer into the program.
    \item \textit{Online Stage}. 
    During the online profiling, the profiler uses the current DL model hyperparameters tuned by CrowdHMTware's elastic inference component (\secref{sec:elastic_model}), to directly calculate the computation load $C$ and parameter size $M$. 
    The accuracy metric $A$ uses the inference confidence proxy without live data labels.
    The profiler also computes dynamic arithmetic intensity $\delta$ for parameters and activations and tests the runtime cache-hit-rate $\epsilon$ (further tuned by CrowdHMTware's co-designed engine in \secref{sec_engine}). 
    These inputs are then used to predict energy demand $E$ and latency $T$ with estimation models for optimization (we will discuss next).

\end{itemize}
\rev{Building on the latency prediction above, we apply the model to the partitioned blocks across devices and include the transmission delay, calculated as the feature size divided by the network bandwidth.}
We note that the primary goal of the above profilers is to ensure \textit{consistent ranking} between the estimated and actual performance on a mobile device.
This consistency is sufficient to provide accurate feedback for automated adjustment.

\rev{\sysname utilizes quick offline measurements to estimate energy and latency during runtime, making it applicable to various devices with different hardware properties.
We estimate the unit transmission latency and computation latency using device memory bandwidth and the processor's FLOPS. 
This allows for accurate estimations across devices without needing runtime data.
Additionally, we utilize offline energy measurement techniques, including the use of the power monitor, to gather ground truth data and calibrate our estimations. 
This ensures that the impact of device heterogeneity is mitigated through pre-measured, device-specific profiles.
This method of latency and energy estimation based on actual device parameters and offline measurement results guarantees the accuracy of the estimations.}

\begin{figure*}[t]
	% \centering
	\begin{subfigure}{0.3\linewidth}
		\centering
		\includegraphics[width=0.98\linewidth]{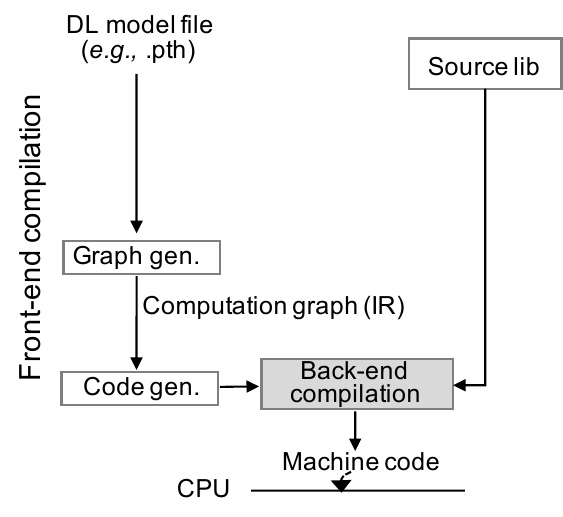}
            \caption{Normal}
		\label{fig:pipeline_a}%文中引用该图片代号
	\end{subfigure}
        \hspace{6mm}
	\centering
	\begin{subfigure}{0.4\linewidth}
		\centering
\includegraphics[width=0.98\linewidth]{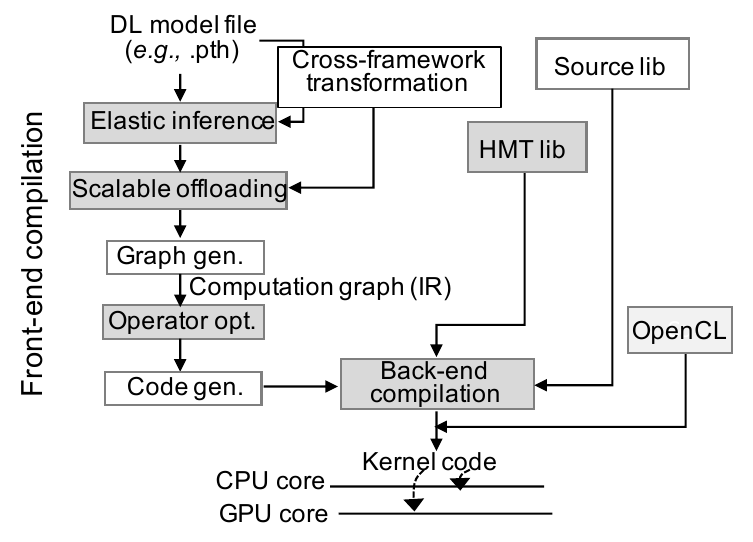}
            \caption{CrowdHMTware}
		\label{fig:pipeline_b}%文中引用该图片代号
	\end{subfigure}
        % \vspace{-2mm}
	\caption{Illustration of the (a) normal and (b) CrowdHMTware front-/back-end compilation routing scheme.}
        \vspace{-2mm}
	\label{fig:pipeline}
\end{figure*}

\subsubsection{Optimizer for Online Adaptation}
\label{subsubsec:optimizer}
CrowdHMTware uses a runtime heuristic optimizer for the constrained multi-objective optimization of accuracy $A$, memory usage $M$, latency $T$, and energy cost $E$. 
This is formulated as the following time-varying optimization problem:
\begin{eqnarray}\label{equ_opt}
\mathop{arg min} \limits_{\theta, \xi} & \mu  Norm(A)  - (1-\mu ) Norm(E) \nonumber \\
\text{s.t.} & T(t)\leq T_{bgt}(t) , \,\, M(t) \leq M_{bgt}(t)
\label{equ:opt}
\end{eqnarray}
where accuracy $A$ and energy cost $E$ are balanced using coefficients $\mu$ and $(1-\mu)$.
$\theta_p$, $\theta_o$, $\theta_s$ represent the tunable DL model hyperparameters (compression operators $\eta_1 \sim \eta_6$) and offloading strategies managed by CrowdHMTware’s elastic inference component and operator/memory scheduling components in CrowdHMTware ’s engine, respectively. 
The value of $\mu$ dynamically depends on the platform’s remaining battery $B\_r$ : $\mu =Norm\left( B\_r \right)$.
Norm(.) is a normalization operation for objective aggregation, \textit{e.g.} log(.).

The coefficient, along with the budgets for responsiveness demands $T_{bgt}$ and memory usage budgets $M_{bgt}$, are specified according to diverse user demands and device-imposed resource availability.
The DL task requirements, user demands, and device constraints directly guide the automated \textit{attributes} and \textit{actions} of the middleware components.
Specifically, $\theta_p$, $\theta_o$, and $\theta_s$ represent the tunable DL model hyperparameters and offloading strategies managed by CrowdHMTware's elastic inference component and operator/memory scheduling components in CrowdHMTware's engine, respectively.
% Specifically, $\theta_p, \theta_o, and \theta_s$ represent the tunable DL model hyperparameters and offloading strategies acted by CrowdHMTware's elastic inference component and operator/memory scheduling components in CrowdHMTware's engine, respectively.

We address this runtime optimization problem in two stages.
In the offline stage, the problem is approached as a static issue, where an evolutionary algorithm is employed to explore a broad set of solutions, ultimately identifying the Pareto-optimal front. 
This front is established by ranking diverse model and system configurations based on pre-tested accuracy and energy costs, ensuring historical rankings align with actual performance on mobile devices. 
To enhance the diversity of candidate solutions, we inject channel-wise variance and Gaussian noise into the solutions, based on the importance of trained architectures. 
We consciously avoid assigning importance coefficients to optimization objectives when defining the Pareto front to maintain an unbiased selection of solutions~\cite{farina2004dynamic}.
In the online stage, the decision variables ($\theta_p$, $\theta_o$, $\theta_s$) adjust dynamically according to current system states, whereas the objective space, defined by accuracy ($A$) and energy ($E$), remains constant. 
We utilize an analytical hierarchy process to dynamically assign importance coefficients $\lambda$ to different criteria, which influence the weighting of each solution’s performance. 
The total score is calculated as $\mu A - (1-\mu) E$, aiding in the selection of the most suitable solution under current conditions. 
% For example, if resource demands of the DL model exceed what is available, the optimizer recalibrates the dynamic importance coefficients to select an optimal configuration from the Pareto frontier based on real-time needs.

\subsubsection{Middleware Implementation}
\label{subsec:implementation}
We implement CrowdHMTware as a dynamic, context-adaptive DL deployment middleware designed for heterogeneous mobile devices.
It automatically adjusts DL models, offloading configurations, and system scheduling to optimize performance in real-time.
As a middleware solution, CrowdHMTware effectively manages application diversity, variable resource availability, and complex system requirements. 
It handles various performance constraints (\eg such as accuracy, memory usage, latency, and energy cost), allowing developers to concentrate on core functionalities without needing to optimize DL model structures, tune parameters for data efficiency, or monitor and schedule system resources.
This capability enables developers to focus on primary application features while CrowdHMTware dynamically ensures optimal system performance.

Specifically, beyond the normal compilation routing (\figref{fig:pipeline_a}), CrowdHMTware realizes cross-framework transformation in elastic inference and scalable offloading components ( \figref{fig:pipeline_b}).
It incorporates operator fusion and parallelism optimization before code generation. 
And it includes a HMT library that packages operator fusion and parallelism techniques not supported by the source library, along with providing an API.
Additionally, CrowdHMTware invokes OpenCL~\cite{stone2010opencl} to generate kernel code that partitions operators and allocates resources to facilitate parallel processing across heterogeneous processors/cores.
We implement CrowdHMTware in Python 3.8, integrating the elastic DL inference component, scalable DL offloading component, and model-adaptive compilation engine into a unified external function interface. 
We use the \textit{run.py} interface ($device_{id}$, $model$, $IP$, $PORT$, $fuse$, $quan$) to access CrowdHMTware.
In addition to the source library from the Pytorch framework, we implement a \textit{HMT} library, which includes APIs for the engine. 
During parallel model offloading, heterogeneous devices communicate via device \textit{IP} and \textit{PORT}.

\section{Experiment}
\label{sec:evaluation}

This subsection presents the evaluations of \rev{CrowdHMTware} in terms of various system performance metrics.

\subsection{Experimental Setups}

\textbf{Mobile and Embedded Devices}.
\rev{We test \rev{CrowdHMTware} on a total of 15 devices, including 12 types of mobile devices and 3 types of embedded devices. }
They represent different computing capabilities and heterogeneous processors and verify the versatility of \rev{CrowdHMTware} as middleware.
\textbf{Tasks, Datasets, and Models}.
We evaluate the performance of \rev{CrowdHMTware} on four types of mobile applications.
Mobile acoustic event awareness uses UbiSound~\cite{sicong2017ubiear}, image classification uses Cifar-100~\cite{krizhevsky2009learning} and ImageNet~\cite{deng2009imagenet} datasets, mobile human activity recognition uses the Har~\cite{anguita2013public}, and driver behavior prediction uses  StateFarm~\cite{farm2016state}.
The models involved in the experiments include ResNet18, ResNet34, VGG16, and MobileNetV2.

\textbf{Baselines}. We adopt four categories of typical baselines.
We employ three hand-crafted DL model compression methods that set high standards for \rev{CrowdHMTware} in terms of accuracy, latency, and resource cost, and four on-demand compression/partition methods to verify \rev{CrowdHMTware}'s performance.

\begin{itemize}
    \item \textbf{Handcrafted model compression}: 
    \begin{itemize}
        \item \textbf{Fire}~\cite{iandola2016squeezenet} compose of a $1\times{1}$ conv layer and a conv layer with a mix of $1\times{1}$ and $3\times{3}$ conv filters. It decreases the sizes of input channels and filters.
        \item \textbf{Singular Value Decomposition (SVD)}~\cite{lane2016deepx} decomposes layer to reduce parameters by retaining important singular values and corresponding eigenvectors.
        \item \textbf{MobileNetV2}~\cite{sandler2018mobilenetv2} inverts a residual layer to expand the component before depth-wise convolution.
    \end{itemize}
    \item \textbf{On-demand model compression}: \textbf{AdaDeep(AdaD)}~\cite{liu2020adadeep}
     automatically combines compression techniques to balance accuracy and efficiency.
    \item \textbf{Once-for-all (OFA)}~\cite{cai2019once}: adaptively selects subnetworks from a supernetwork that supports multiple architectures.
    \item \textbf{Adaptive DL model partition}:
    \begin{itemize}
    \item \textbf{CAS}~\cite{wang2021context} heuristically guides model partition for effective adaptation in dynamic runtime contexts.
    \item \textbf{DADS}~\cite{hu2019dynamic} partitions DL models with a DAG topology as a minimum cut problem in graph theory.
    \end{itemize}
    \item \textbf{\rev{CrowdHMTware}} integrates the cross-level adaptive optimization, including hardware-aware algorithm (\ie elastic inference and partition), and model-adaptive engine (\ie operator/resource scheduling) levels, forming an automated adaptation loop.
\end{itemize}

\begin{figure*}[t]
	% \centering
	\begin{subfigure}{0.23\linewidth}
		\centering
		\includegraphics[width=0.90\linewidth]{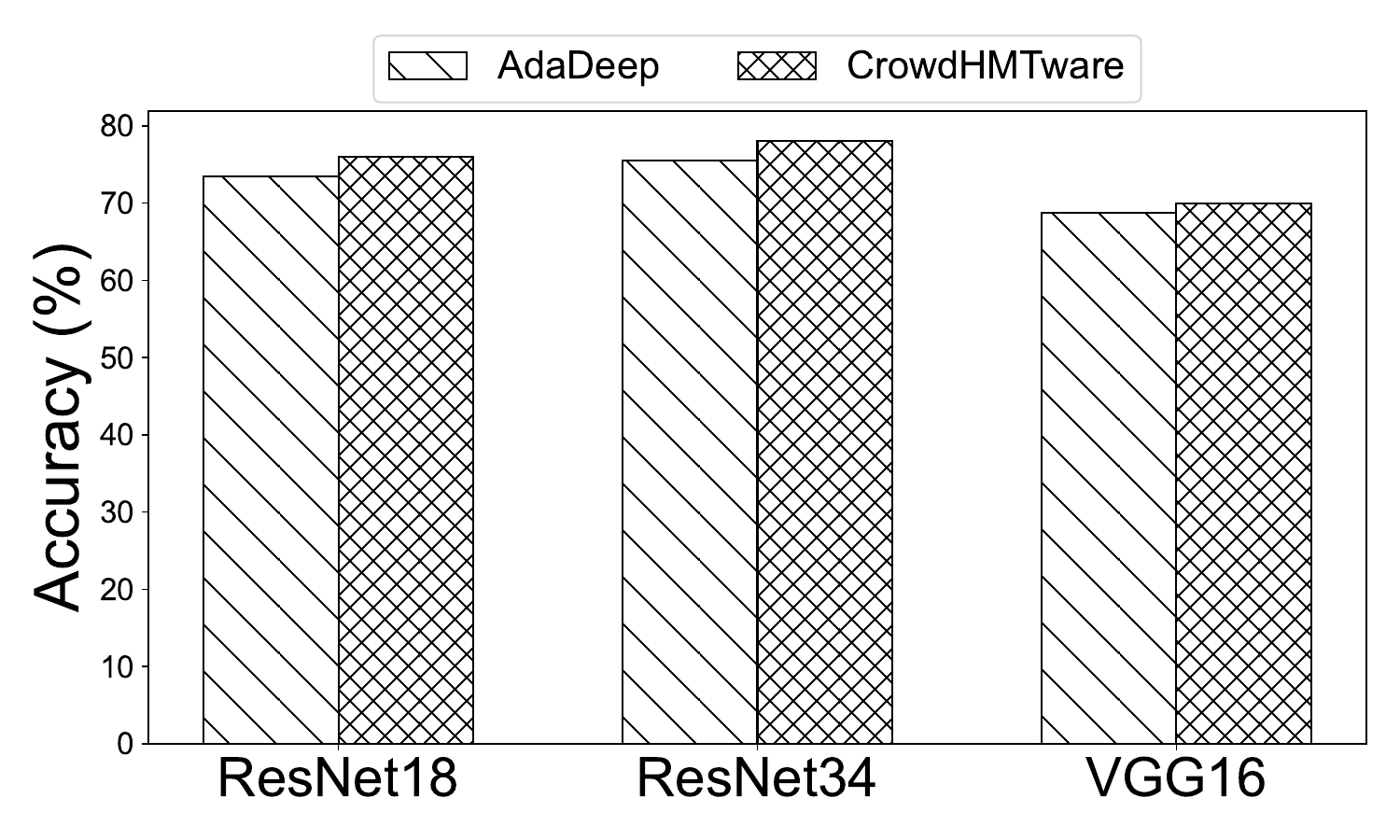}
            \caption{Accuracy}
		\label{fig:model_acc}%文中引用该图片代号
	\end{subfigure}
	% \centering
	\begin{subfigure}{0.23\linewidth}
		\centering
		\includegraphics[width=0.90\linewidth]{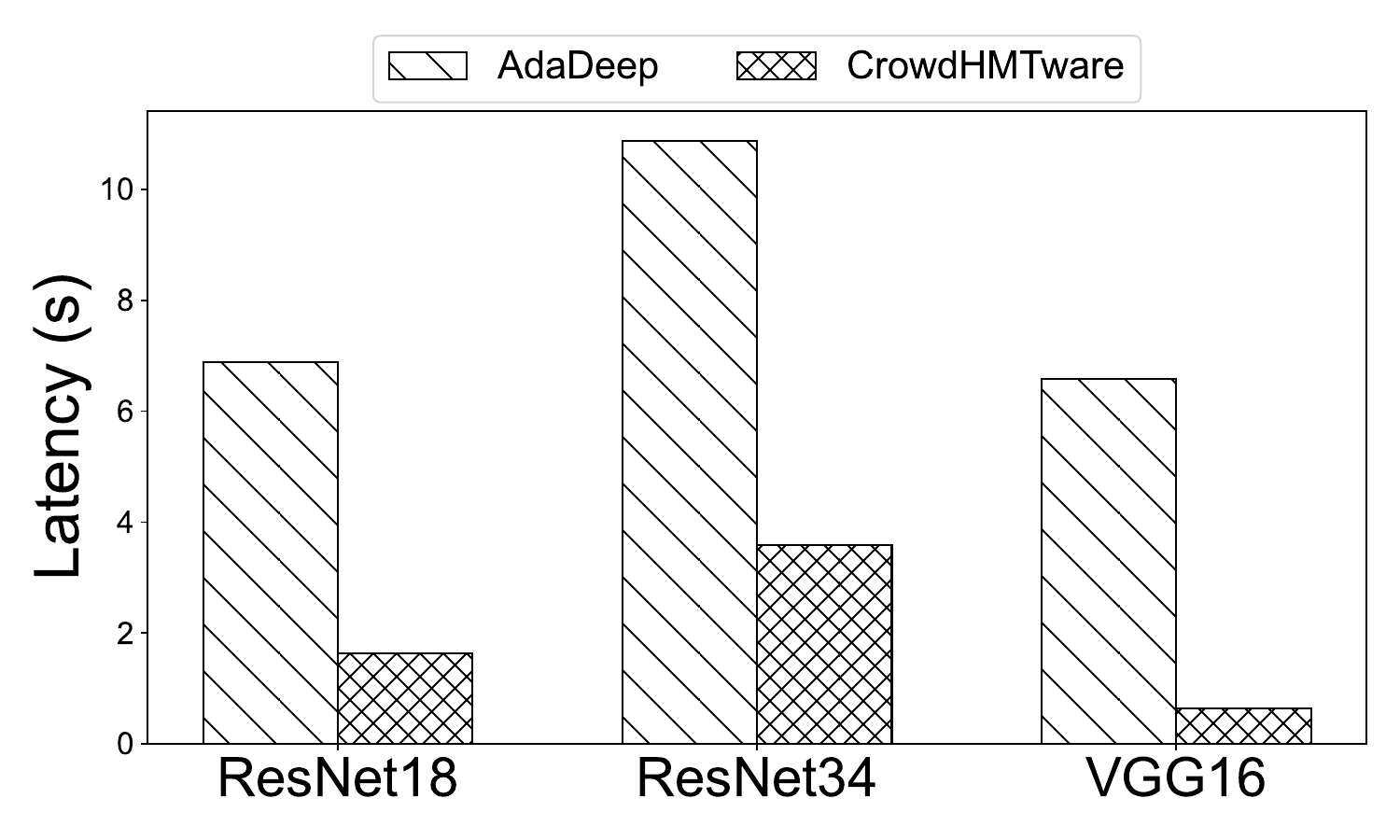}
            \caption{Latency}
		\label{fig:model_latency}%文中引用该图片代号
	\end{subfigure}
        % \centering
	\begin{subfigure}{0.23\linewidth}
		\centering
		\includegraphics[width=0.90\linewidth]{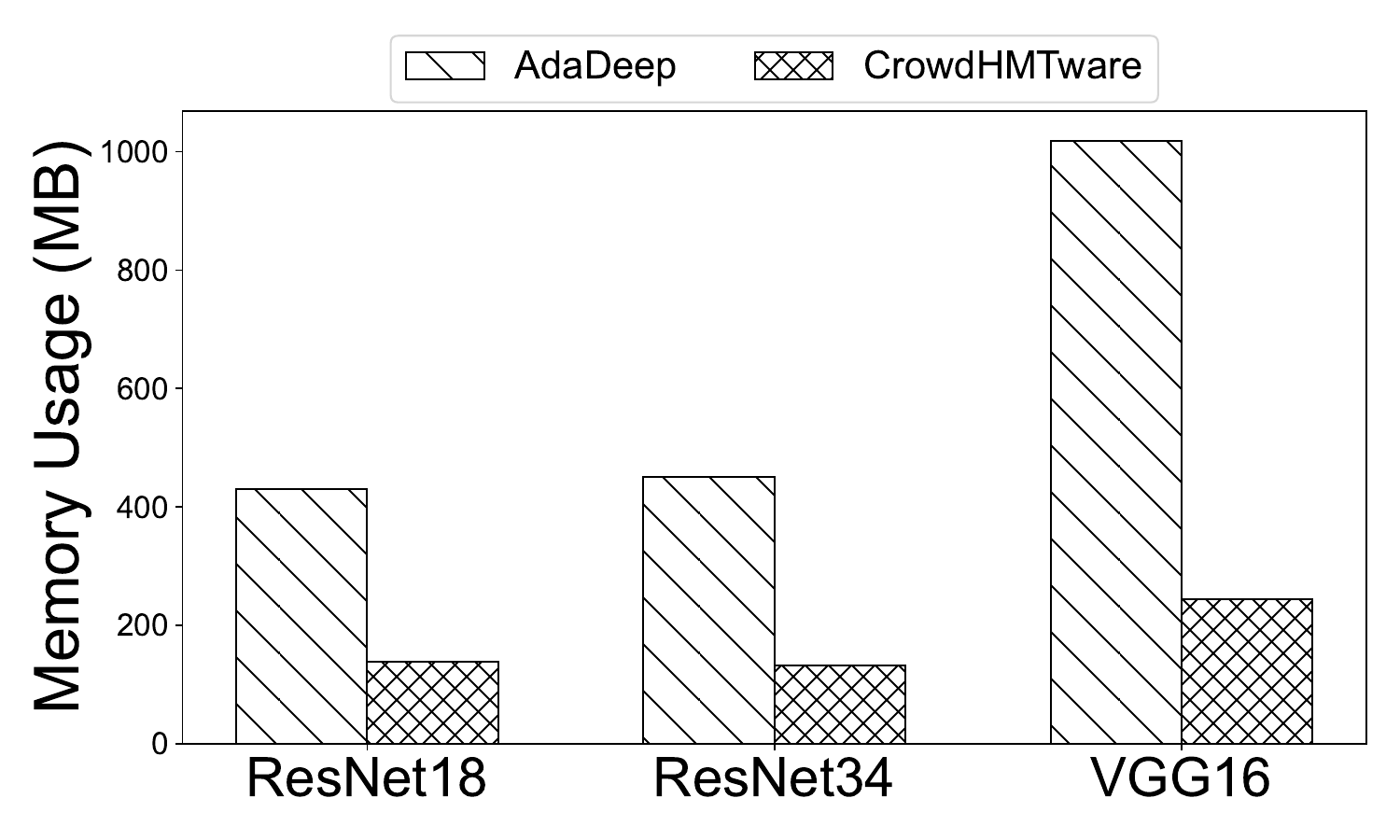}
            \caption{Memory usage}
		\label{fig:model_mem}%文中引用该图片代号
	\end{subfigure}
        % \centering
	\begin{subfigure}{0.23\linewidth}
		\centering
		\includegraphics[width=0.90\linewidth]{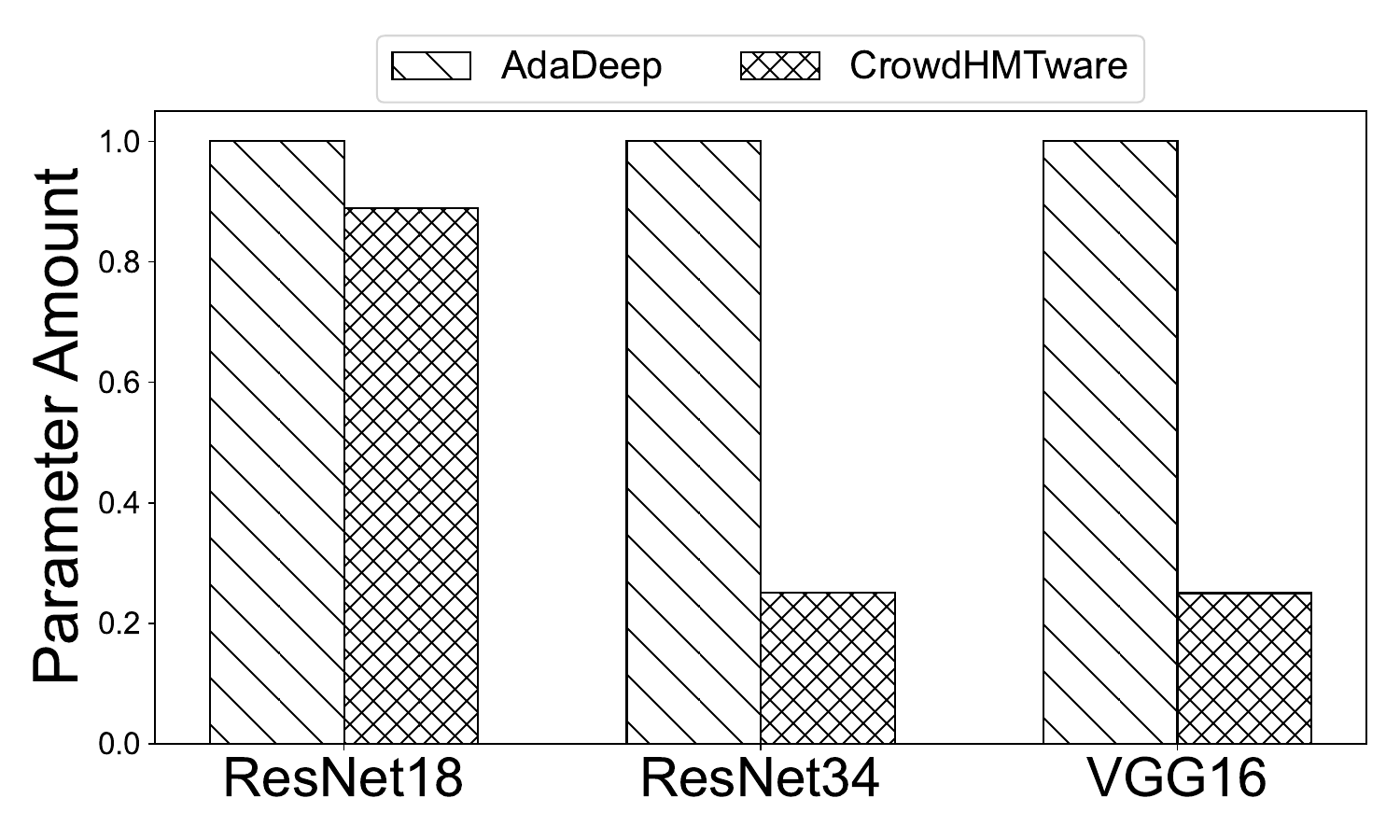}
            \caption{Parameter Amount}
		\label{fig:model_param}%文中引用该图片代号
	\end{subfigure}
        \vspace{-2mm}
	\caption{Performance comparison of \rev{CrowdHMTware} with AdaDeep over ResNet18, ResNet34, VGG16 models.}
        \vspace{-2mm}
	\label{fig:model}
\end{figure*}

\subsection{Overall Performance}
We compare \rev{CrowdHMTware}'s performance with baselines across different models, devices, and user demands.

\subsubsection{Performance Comparison over Diverse DL models}
We evaluate the performance of \rev{CrowdHMTware} using ResNet18, ResNet34, and VGG16 on a Raspberry Pi 4B. 
The results, compared with AdaDeep, are depicted in Fig.~\ref{fig:model}. 
\textit{First}, \rev{CrowdHMTware} achieves the best overall trade-off due to cross-level optimization spanning models, offloading, operations, and underlying resources, which enhanced single-level performance bounds. 
\textit{Second}, \rev{CrowdHMTware} also delivered higher accuracy, as the underlying engine provided greater flexibility for higher-level models. 
\textit{Third}, \rev{CrowdHMTware} significantly reduces latency. 
\rev{CrowdHMTware}'s latency is 4.2$\times$ lower on ResNet18, 3$\times$ lower on ResNet34, and 10.3$\times$ lower on VGG16. 
For memory usage, \rev{CrowdHMTware} is 3.1$\times$, 3.4$\times$, and 4.2$\times$ lower than AdaDeep on ResNet18, ResNet34, and VGG16.
% The parameter amount of \rev{CrowdHMTware} is reduced by 1.1$\times$-4$\times$.

\begin{figure}[t]
	\centering
	\begin{subfigure}{0.6\linewidth}
		\centering		\includegraphics[width=0.9\linewidth]{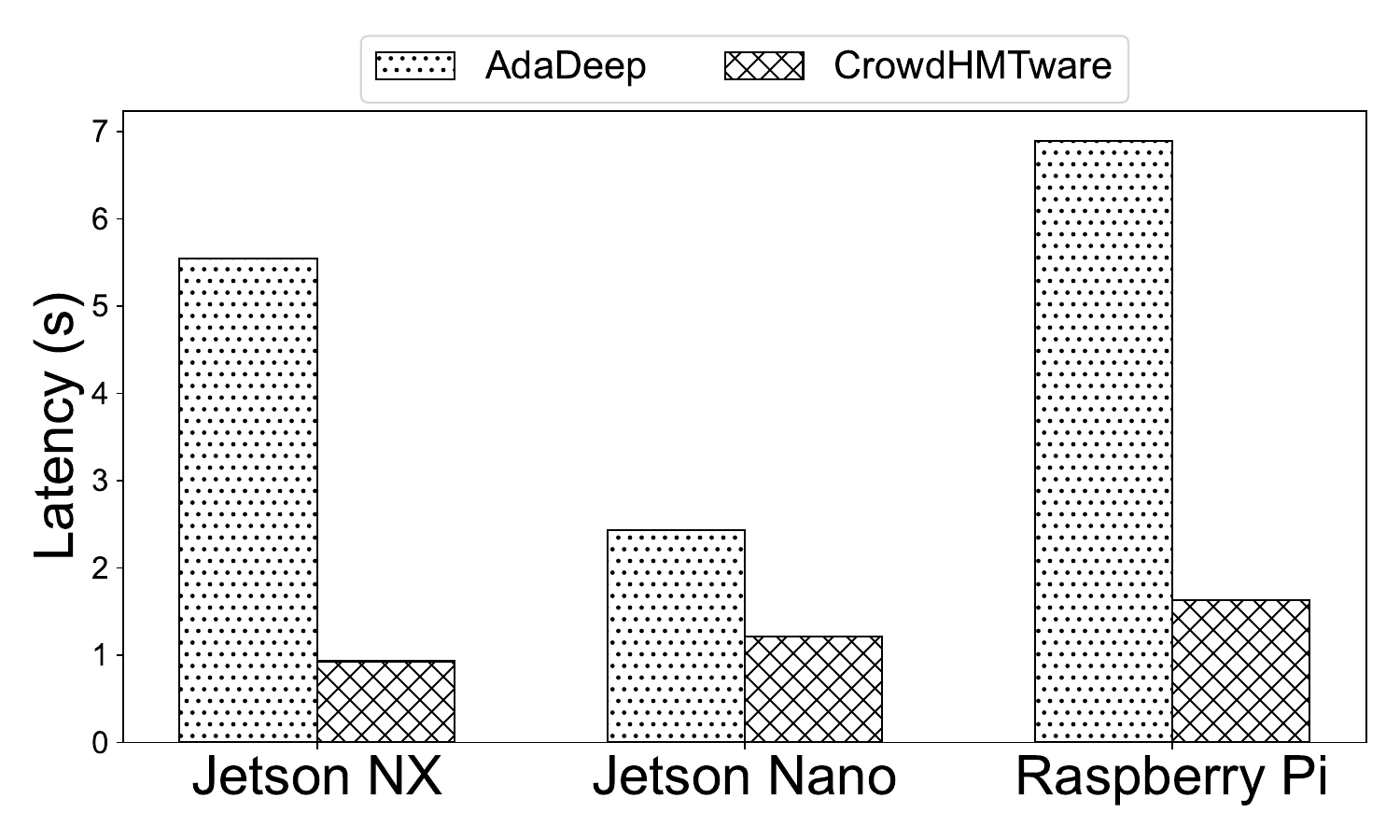}
            \caption{Latency}
		\label{fig:device_latency}%文中引用该图片代号
	\end{subfigure}
	\begin{subfigure}{0.75\linewidth}
	\centering
 \includegraphics[width=0.7\linewidth]{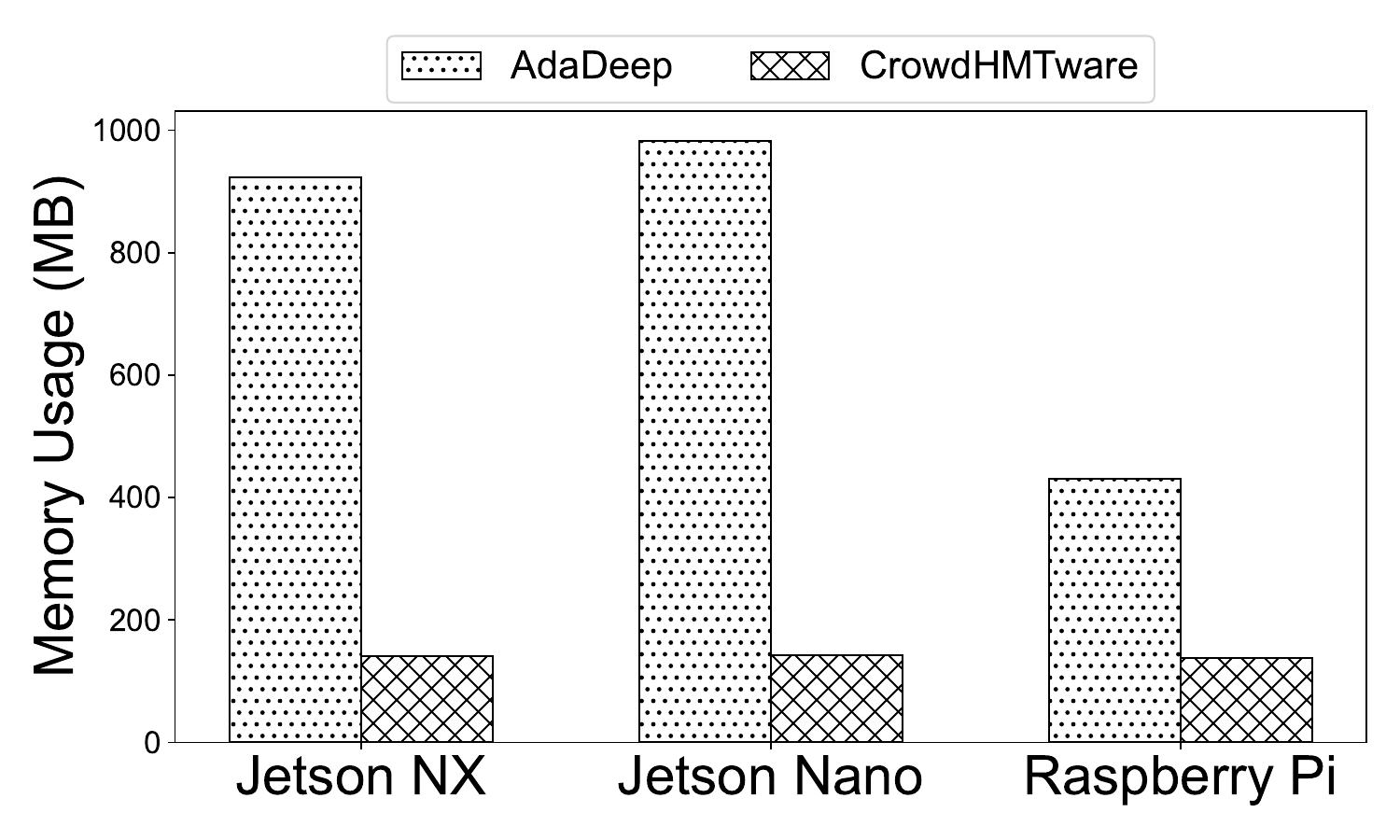}
            \caption{Memory usage}
		\label{fig:device_mem}%文中引用该图片代号
	\end{subfigure}
        \vspace{-1mm}
	\caption{Performance comparison of \rev{CrowdHMTware} with AdaDeep across different devices.}
         \vspace{-1mm}
	\label{fig:device}
\end{figure}

\subsubsection{Performance Comparison on Heterogeneous Mobile Devices}
We evaluate \rev{CrowdHMTware}'s performance using ResNet18 for inference on Jetson NX, Jetson Nano, and Raspberry Pi 4B, with results compared to AdaDeep shown in ~\figref{fig:device}. 
On the Raspberry Pi 4B, \rev{CrowdHMTware} achieves a 4.2$\times$ reduction in latency and a 3.1$\times$ reduction in memory usage. 
Similar improvements are observed on both Jetson NX and Jetson Nano.
We also compare \rev{CrowdHMTware} with the original model on 12 popular mobile and embedded platforms, which include smartphones, wearable devices, development boards, and smart home devices, all equipped with different processors, storage capacities, and batteries.
Performance such as accuracy, latency, and energy cost have all been improved to varying degrees shown in Table~\ref{table:diverse_device}.

\begin{table}[]
\centering
\scriptsize
\caption{Performance of \rev{CrowdHMTware} on 12 mobile and embedded devices normalized by the original model.}
\vspace{-1mm}
\label{table:diverse_device}
\renewcommand{\arraystretch}{1.1}
\begin{tabular}{|c|c|c|c|c|}
\hline
\textbf{Mobile $\&$ embedded device}          & \textbf{Accuracy} & \textbf{Latency} & \textbf{MACs} & \textbf{Energy} \\ \hline
\textbf{Samsung note5}   & 1.20\%            & 1.6$\times$             & 4.1$\times$          & 1.8$\times$            \\ \hline
\textbf{Huawei iP9}      & 0.90\%            & 1.6$\times$             & 1.6$\times$          & 1.7$\times$            \\ \hline
\textbf{Huawei pra-a100} & 1.20\%            & 1.4$\times$             & 6.8$\times$          & 1.8$\times$            \\ \hline
\textbf{Xiaomi Mi 6}     & 1.10\%            & 1.9$\times$             & 5.6$\times$          & 1.6$\times$            \\ \hline
\textbf{Xiaomi Mi 5S}    & 1.80\%            & 2.1$\times$             & 3.6$\times$          & 1.2$\times$            \\ \hline
\textbf{Xiaomi Redmi 3S} & 0.90\%            & 1.6$\times$             & 2.1$\times$          & 1.1$\times$            \\ \hline
\textbf{Huawei watchH2P} & 2.10\%            & 3.1$\times$             & 3.6$\times$          & 8.3$\times$            \\ \hline
\textbf{Sony watch $SW_3$}  & 1.60\%            & 1.5$\times$             & 2.1$\times$          & 9.8$\times$            \\ \hline
\textbf{firefly-rk3999}  & 1.80\%            & 2.6$\times$             & 5.6$\times$          & 1.2$\times$            \\ \hline
\textbf{firefly-rk3288}  & 0.70\%            & 1.1$\times$             & 4.8$\times$          & 1.3$\times$            \\ \hline
\textbf{Huawei box}      & 1.90\%            & 2.8$\times$             & 1.6$\times$          & 1.2$\times$            \\ \hline
\textbf{Xiaomi box 3S}   & 1.20\%            & 1.4$\times$             & 4.1$\times$          & 1.1$\times$            \\ \hline
\end{tabular}
% \vspace{-3mm}
\end{table}

\begin{table*}[]
\centering
\small
\caption{Performance comparison of \rev{CrowdHMTware} under different resource budgets.}
% \vspace{-1mm}
\label{table:dynamic}
\begin{tabular}{|c|cccc|}
\hline
\multirow{2}{*}{\textbf{Performance}} & \multicolumn{4}{c|}{\textbf{Resource usage(MB)}}                                                                                                                               \\ \cline{2-5} 
                                      & \multicolumn{1}{c|}{\textbf{Non-Restriction}} & \multicolumn{1}{c|}{\textbf{75\%Memory Budget}} & \multicolumn{1}{c|}{\textbf{50\%Memory Budget}} & \textbf{25\%Memory Budget} \\ \hline
\textbf{Accuracy (\%)}                 & \multicolumn{1}{c|}{75}                       & \multicolumn{1}{c|}{76}                      & \multicolumn{1}{c|}{76}                      & 76                      \\ \hline
\textbf{Latency (s)}                   & \multicolumn{1}{c|}{6.93}                     & \multicolumn{1}{c|}{5.58}                       & \multicolumn{1}{c|}{3.03}                       & 11.04                      \\ \hline
\textbf{Memory (MB)}                   & \multicolumn{1}{c|}{699.08}                   & \multicolumn{1}{c|}{524.84}                     & \multicolumn{1}{c|}{354.38}                     & 127.29                     \\ \hline
\end{tabular}
\vspace{-1mm}
\end{table*}

\subsubsection{Performance Comparison under Dynamic Resource Budgets}
We compare the adaptive performance of \rev{CrowdHMTware} under different resource budgets using ResNet18 on Raspberry Pi 4B, as shown in Table~\ref{table:dynamic}.
We simulate dynamic context in real-world deployment scenarios by setting different memory resource limits. 
\textit{First}, 
when the memory budget is 75\%, \rev{CrowdHMTware}'s memory usages drop 25\% compared to the non-restriction state.
And latency is 1.24$\times$ lower than the non-restriction state. 
\textit{Second}, with 50\% memory budget, \rev{CrowdHMTware} further reduces its memory usage to 354.38MB with a 56\% reduction in latency. 
\textit{Third}, in the extremely constrained state with $25\%$ memory budget, the memory usage of \rev{CrowdHMTware} reduces to 127.29MB, which is 18\% of the non-restriction case.
Meanwhile, the accuracy is still maintaining, and the extreme state caused the latency to increase.
Thanks to its cross-level optimization through elastic model scaling, offloading, and memory scheduling, \rev{CrowdHMTware} effectively optimizes memory usage while maintaining accuracy and minimizing latency increases.

\begin{figure*}[t]
	% \centering
	\begin{subfigure}{0.19\linewidth}
		\centering
		\includegraphics[width=0.95\linewidth]{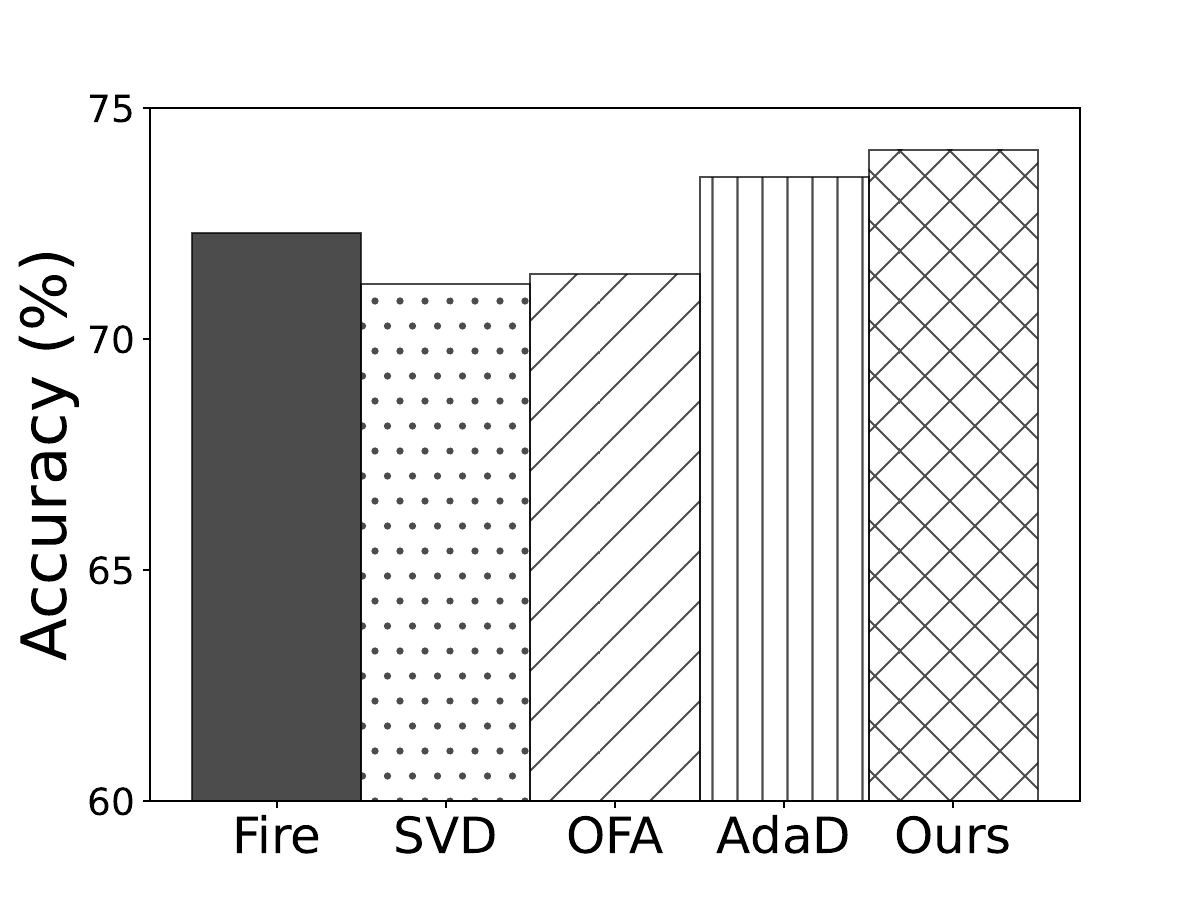}
            \caption{Accuracy}
		\label{fig:comp_acc}%文中引用该图片代号
	\end{subfigure}
	% \centering
	\begin{subfigure}{0.19\linewidth}
		\centering
		\includegraphics[width=0.95\linewidth]{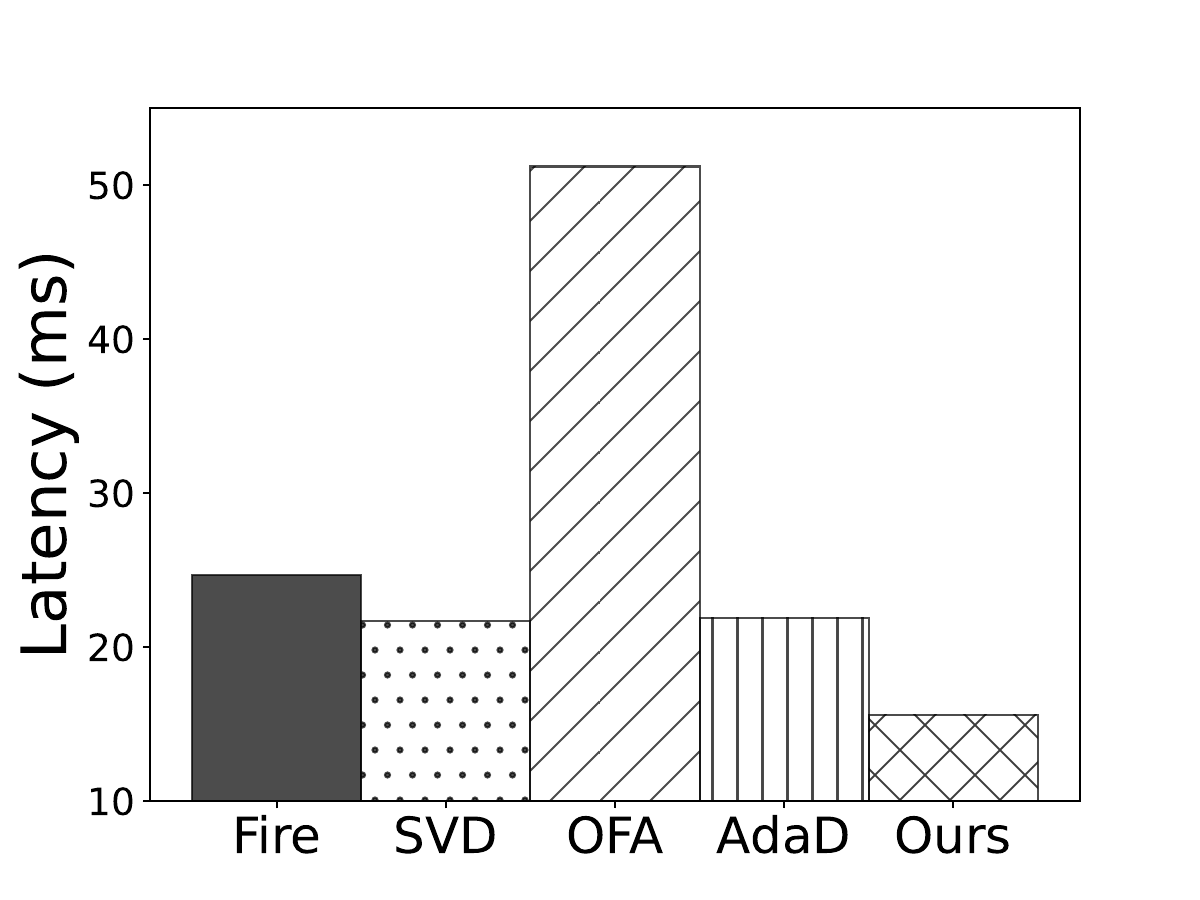}
            \caption{Latency}
		\label{fig:comp_latency}%文中引用该图片代号
	\end{subfigure}
	% \centering
	\begin{subfigure}{0.19\linewidth}
		\centering
		\includegraphics[width=0.95\linewidth]{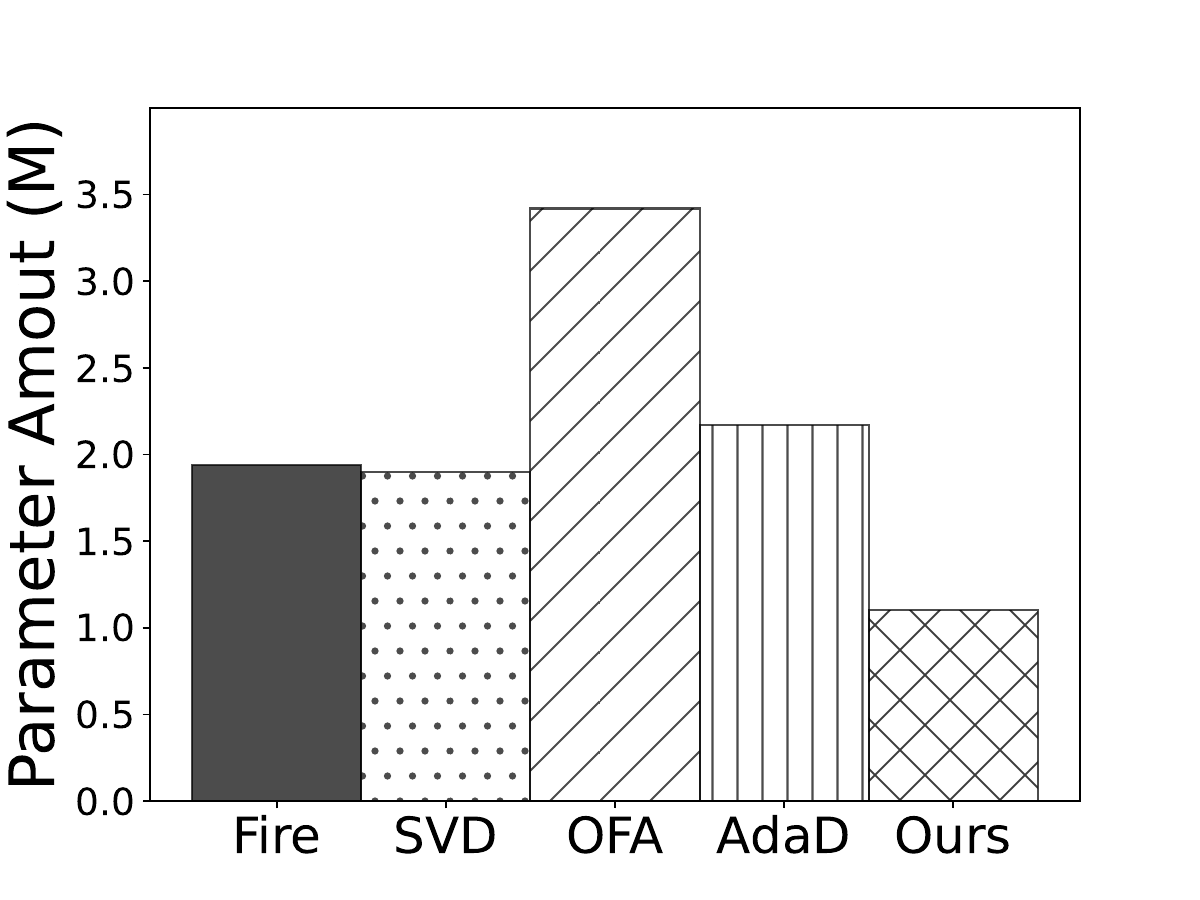}
            \caption{Parameter Amount}
		\label{fig:comp_param}%文中引用该图片代号
	\end{subfigure}
        % \centering
	\begin{subfigure}{0.19\linewidth}
		\centering
		\includegraphics[width=0.95\linewidth]{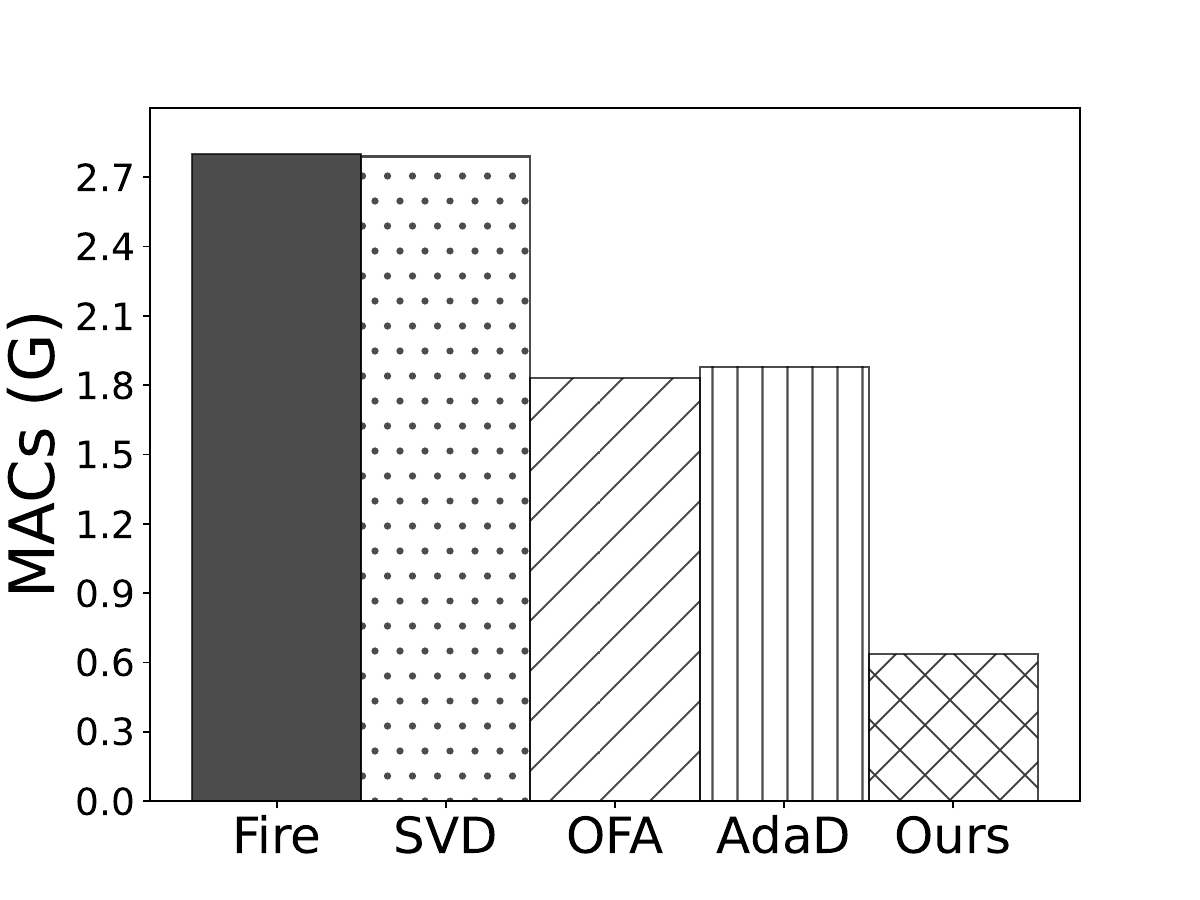}
            \caption{MACs}
		\label{fig:comp_mac}%文中引用该图片代号
	\end{subfigure}
        % \centering
	\begin{subfigure}{0.19\linewidth}
		\centering
		\includegraphics[width=0.95\linewidth]{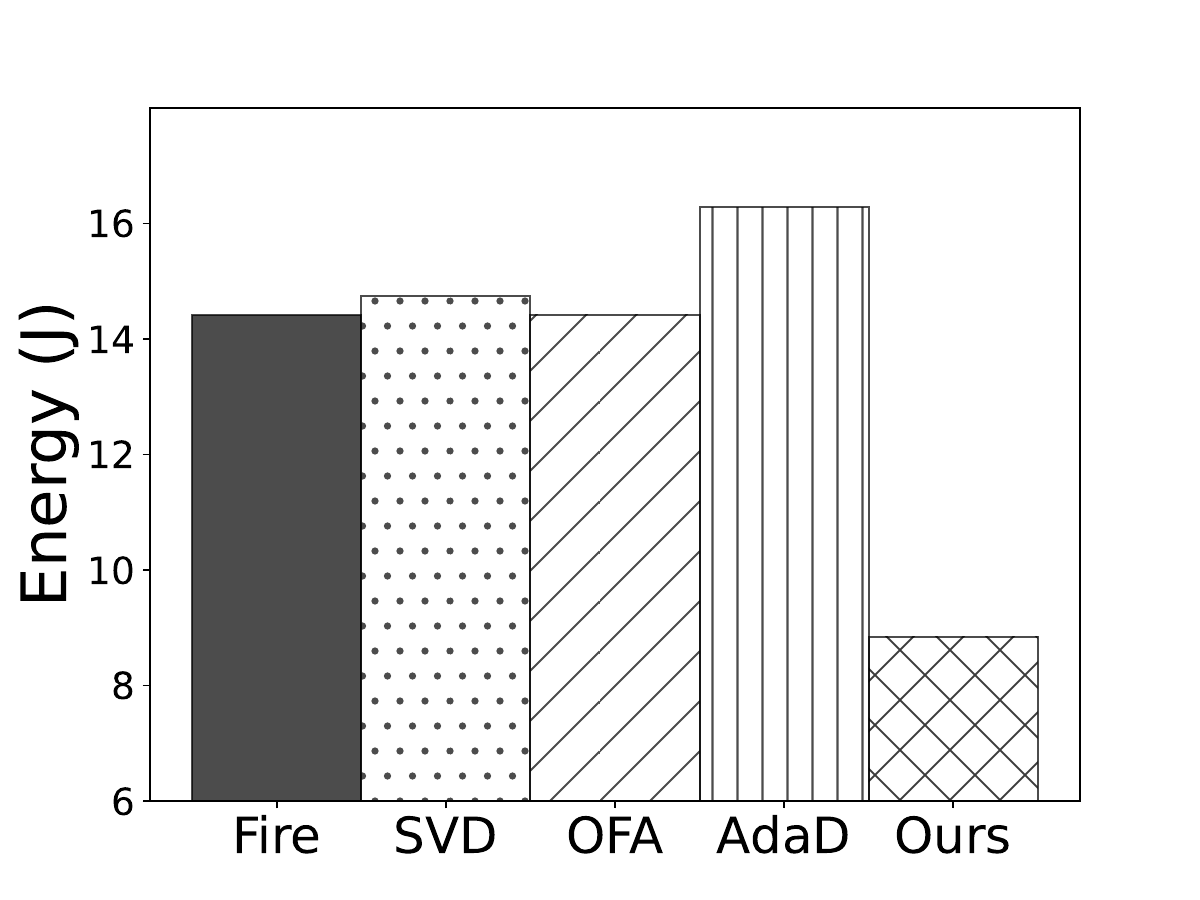}
            \caption{Energy}
		\label{fig:comp_energy}%文中引用该图片代号
	\end{subfigure}
        \vspace{-1mm}
	\caption{Performance comparison of \rev{CrowdHMTware}'s elastic inference component against baselines.}
        % \vspace{-3mm}
	\label{fig:compression}
\end{figure*}

\begin{table}[t]
\centering
\tiny
\caption{Performance comparison using different compression operator combinations on different tasks/datasets.}
\vspace{-1mm}
\label{table:operator_combine}
\renewcommand{\arraystretch}{1.3}
\scalebox{1.2}{
\begin{tabular}{|c|ccccc|}
\hline
\multirow{2}{*}{\textbf{\begin{tabular}[c]{@{}c@{}}Operator\\ Combination\end{tabular}}} & \multicolumn{5}{c|}{\textbf{Compared to MobileNetV2 baseline}}                                                                                                                          \\ \cline{2-6} 
                                                                                                       & \multicolumn{1}{c|}{\textbf{Accuracy}} & \multicolumn{1}{c|}{\textbf{Latency}} & \multicolumn{1}{c|}{\textbf{MAC}} & \multicolumn{1}{c|}{\textbf{Energy}} & \textbf{Dataset}   \\ \hline
\textbf{$\eta_1$ + $\eta_6$}                                                                                         & \multicolumn{1}{c|}{1.30\%}            & \multicolumn{1}{c|}{1.1$\times$}             & \multicolumn{1}{c|}{4.3$\times$}         & \multicolumn{1}{c|}{15.2$\times$}           & \textbf{UbiSound~\cite{sicong2017ubiear}}  \\ \hline
\textbf{$\eta_2$ + $\eta_6$}                                                                                         & \multicolumn{1}{c|}{-2.10\%}           & \multicolumn{1}{c|}{1.2$\times$}             & \multicolumn{1}{c|}{5.6$\times$}         & \multicolumn{1}{c|}{2.5$\times$}            & \textbf{Cifar-100~\cite{krizhevsky2009learning}} \\ \hline
\textbf{$\eta_1$ + $\eta_5$}                                                                                         & \multicolumn{1}{c|}{-0.90\%}           & \multicolumn{1}{c|}{1.3$\times$}             & \multicolumn{1}{c|}{8.6$\times$}         & \multicolumn{1}{c|}{8.9$\times$}            & \textbf{ImageNet~\cite{deng2009imagenet}}  \\ \hline
\textbf{$\eta_2$ + $\eta_5$}                                                                                         & \multicolumn{1}{c|}{-0.30\%}           & \multicolumn{1}{c|}{0.8$\times$}             & \multicolumn{1}{c|}{9.2$\times$}         & \multicolumn{1}{c|}{2.1$\times$}            & \textbf{Har~\cite{anguita2013public}}       \\ \hline
\textbf{$\eta_1$ + $\eta_6$}                                                                                         & \multicolumn{1}{c|}{0.20\%}            & \multicolumn{1}{c|}{0.7$\times$}             & \multicolumn{1}{c|}{5.6$\times$}         & \multicolumn{1}{c|}{5.9$\times$}            & \textbf{StateFarm~\cite{farm2016state}} \\ \hline
\end{tabular}
}
% \vspace{-4mm}
\end{table}

\subsection{Performance of Elastic Inference Comp.}

We evaluate \rev{CrowdHMTware}'s elastic inference component on the Raspberry Pi 4B using the Cifar-100 dataset, comparing it against model compression baselines like Fire, SVD, Once-for-all, and AdaDeep. 
\textit{First}, the results, detailed in \figref{fig:compression}, demonstrate significant improvements in multiple metrics: accuracy (\figref{fig:comp_acc}), latency (\figref{fig:comp_latency}), parameter efficiency (\figref{fig:comp_param}), MAC operations (\figref{fig:comp_mac}), and energy cost (\figref{fig:comp_energy}).
\textit{Second}, we evaluate \rev{CrowdHMTware}'s performance on other models, where it autonomously selects different compression operators to optimize resources and enhance performance, as detailed in Table \ref{table:operator_combine}. 
Compared to MobileNetV2, \rev{CrowdHMTware} reduces latency by 0.7$\times$ to 1.3$\times$, cut computational complexity by 5.6$\times$ to 9.2$\times$, and decreases energy consumption by 2.1$\times$ to 15.2$\times$ with minimal impact on accuracy.

\begin{figure}[t]
	\centering
	\begin{subfigure}{0.75\linewidth}
		\centering
		\includegraphics[width=0.98\linewidth]{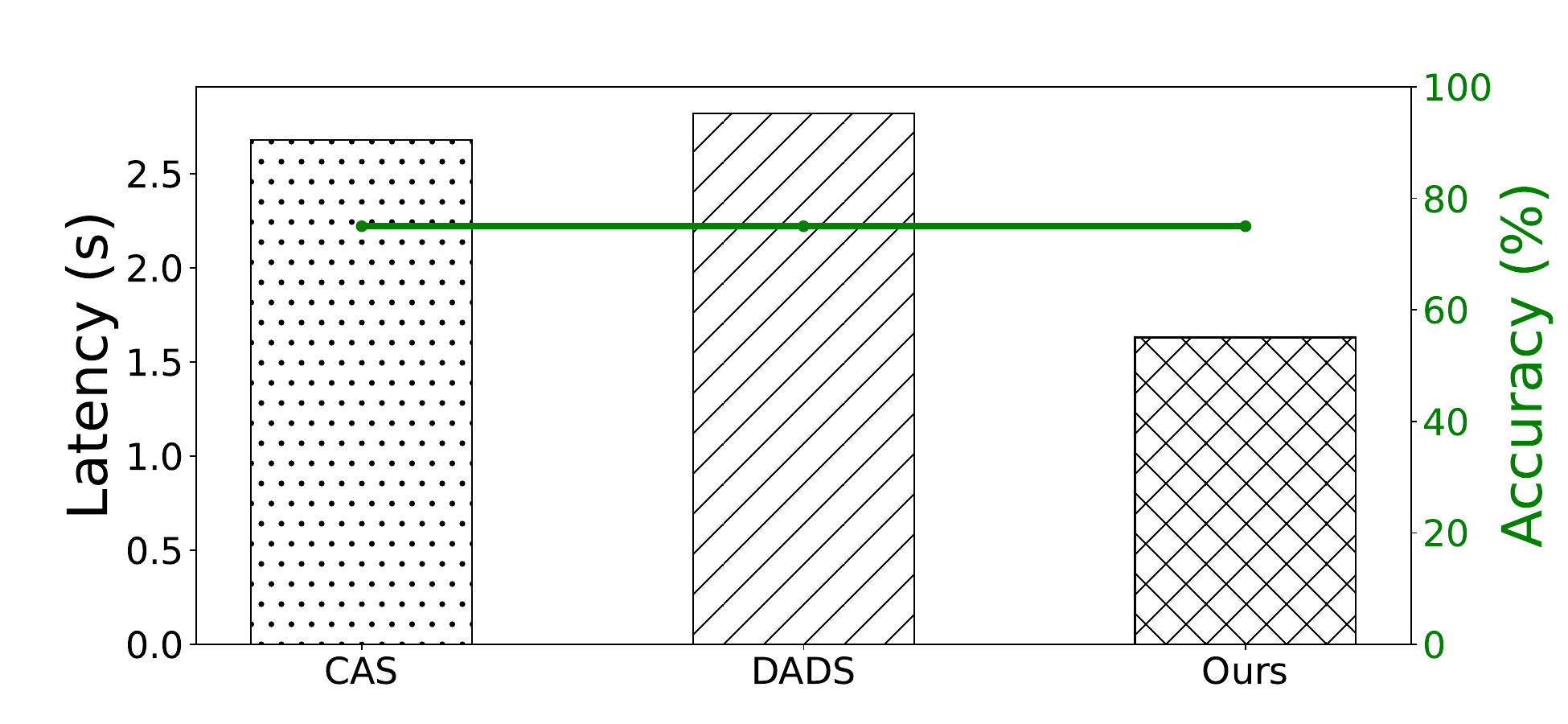}
            \caption{Latency and accuracy.}
	\label{fig:part_latency_acc}%文中引用该图片代号
	\end{subfigure}
	\begin{subfigure}{0.75\linewidth}
		\centering	\includegraphics[width=0.98\linewidth]{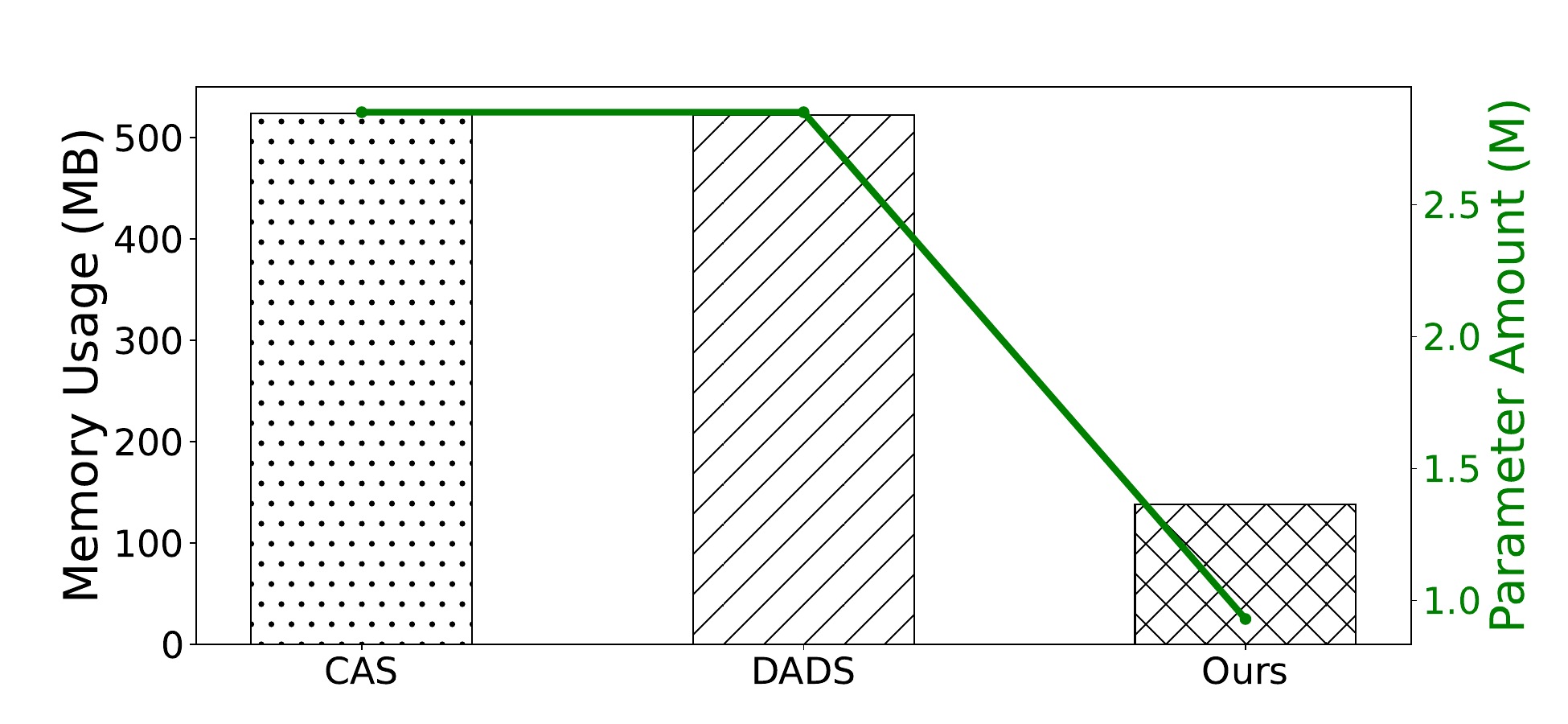}
            \caption{Memory and parameter size.}
		\label{fig:part_mem_param}%文中引用该图片代号
	\end{subfigure}
        \vspace{-1mm}
	\caption{Performance of \rev{CrowdHMTware}'s scalable DL model offloading component compared with baselines.}
        % \vspace{-3mm}
	\label{fig:partition}
\end{figure}

\subsection{Performance of Scalable Offloading Comp.}
We test \rev{CrowdHMTware} with CAS and DADS using ResNet18 on Raspberry Pi 4B, as illustrated in \figref{fig:partition}. 
\textit{First}, \rev{CrowdHMTware} significantly reduces latency by 39\% and 42\% compared to CAS and DADS, respectively, without notably affecting accuracy (\figref{fig:part_latency_acc}). 
\rev{CrowdHMTware} reduces memory usage by 385.5MB, a 74\% decrease compared to CAS, while also reducing the parameter count by 67\%. 
\textit{Second}, against DADS, memory and parameter reductions are 73\% and 67\%, respectively (\figref{fig:part_mem_param}).

\begin{table*}[t]
\centering
\scriptsize
\caption{Performance of cross-level optimization of parallel inference on mobile device with Snapdragon 855.}
\renewcommand{\arraystretch}{1.1}
\vspace{-1mm}
\label{tab:cross}
% \resizebox{1.5\columnwidth}{!}{
\begin{tabular}{|c|c|c|c|c|c|}
\hline
\textbf{Levels} & \textbf{Methods} & \textbf{\begin{tabular}[c]{@{}c@{}}Top\\ accuracy (\%)\end{tabular}} & \textbf{\begin{tabular}[c]{@{}c@{}}Memory \\ usage (MB)\end{tabular}} & \textbf{\begin{tabular}[c]{@{}c@{}}Latency\\ (ms)\end{tabular}} & \textbf{\begin{tabular}[c]{@{}c@{}}Speedup\\ (\%)\end{tabular}} \\ \hline

Original model& ResNet-18~\cite{he2016deep} & 76.23 & 47.24 & 213.24 & --- \\ \hline 

\multirow{2}{*}{\begin{tabular}[c]{@{}c@{}}Resource-friendly\\ frontend compilation\end{tabular}}
 & Low-rank decomposition & 73.73 & 15.61 & 197.53 & 7.37 \\ \cline{2-6} 
 & Pruning & 71.31 & 23.91 & 146.73 & 31.19 \\ \hline
\multirow{2}{*}{\begin{tabular}[c]{@{}c@{}}Model-adaptive\\ backend compilation\end{tabular}} & \begin{tabular}[c]{@{}c@{}}Operator parallelism \\ \end{tabular} & 76.23 & 47.24 & 189.01 & 11.36 \\ \cline{2-6} 
 & \begin{tabular}[c]{@{}c@{}}Operator fusion \end{tabular} & 76.23 & 47.24 & 136.66 & 35.91\\ \hline

\multirow{3}{*}{\begin{tabular}[c]{@{}c@{}}Cross-level \\ optimization\end{tabular}} & \begin{tabular}[c]{@{}c@{}}Operator parallelism+Low-rank decomposition \\ \end{tabular} & 73.72 & 15.60 & 132.96 & 37.65 \\ \cline{2-6} 
 & \begin{tabular}[c]{@{}c@{}}Operator parallelism+ pruning \end{tabular} & 71.31 & 23.89 & 131.46 & 38.35\\
\cline{2-6} 
 & Operator parallelism+pruning+operator fusion+memory allocation & 73.56 & 11.53 & 103.23 & 48.4\% \\ \hline
\end{tabular}
% \vspace{-2mm}
\end{table*}

\subsection{Performance of Model-adaptive Engine}
To assess the dynamic model-adaptive engine for adaptive DL inference, we tested it against various baselines using ResNet-18, detailed in Table~\ref{tab:cross}. 
Initially, frontend compilation optimizations enhance resource utilization, where low-rank decomposition cuts latency by 7.37\% through smaller matrix use, and channel pruning reduces it further by 31.19\% by eliminating redundant channels. 
At the backend level, compilation enhancements like operator fusion and parallelism expand resource availability, with operator parallelism on CPU+GPU improving inference speed by 11\% and fusion reducing latency by 35\%. 
Cross-level optimization effectively balances accuracy and efficiency, combining pruning, fusion, memory allocation, and parallelism to reduce latency by 48.4\%.

\subsection{Ablation Study}
To verify the impact of each component in \rev{CrowdHMTware} on performance improvement, we design an ablation study with five experiments using ResNet18 on Raspberry Pi 4B. 
These include combinations of the elastic compression and partitioning components, elastic compression and engine components, partitioning and engine components, and the full \rev{CrowdHMTware} system. 
Table \ref{table:ablation} presents the results of our ablation study on \rev{CrowdHMTware}.
\textit{First}, integrating scalable offloading component and engine within the elastic inference component significantly reduces latency by 44\% and parameter count by 72\%. 
This reduction in parameters leads to lower computational complexity and, consequently, quicker computation times.
\textit{Second}, the scalable DL model offloading, when compared to the elastic compression + engine, cuts latency by 65\% and memory usage by 74\%. 
By dividing the model into sub-models that infer in parallel, it effectively minimizes both latency and memory overhead without altering the model's structure, thus preserving accuracy and parameter size.
\textit{Third}, versus elastic compression + scalable offloading, \rev{CrowdHMTware} achieves a 32\% reduction in latency, 66\% in memory usage, and 75\% in parameter count through cross-level optimization. 
These improvements are attributed to operator fusion, which decreases memory accesses, advanced memory allocation that minimizes redundancy, and pruning which significantly reduces parameter count, aiding in both memory and latency reductions.

\begin{table*}[t]
\centering
\caption{Impact of each component on the overall performance of \rev{CrowdHMTware}.}
\vspace{-1mm}
\footnotesize
\label{table:ablation}
\renewcommand{\arraystretch}{1.1}
\begin{tabular}{|c|c|c|c|c|}
\hline
                                   \textbf{Method} &
                                    \textbf{Accuracy} & \textbf{Latency} & \textbf{Memory}   & \textbf{Parameter} \\ \hline
\textbf{DL model compression + DL model partitioning}   & 76.00             & 2.39s            & 410.11MB          & 3.19M              \\ \hline
\textbf{DL model  compression + Engine} & 76.00    & 4.72s            & 528.68MB          & \textbf{0.80M}     \\ \hline
\textbf{DL model partitioning + Engine}   & 75.00             & 2.91s            & 142.57MB          & 2.81M              \\ \hline
\textbf{\rev{CrowdHMTware} (DL model compression + partitioning + Engine)}              & 76.00    & \textbf{1.63s}   & \textbf{138.41MB} & \textbf{0.80M}     \\ \hline
\end{tabular}
\end{table*}

\begin{figure*}[t]
	% \centering
	\begin{subfigure}{0.23\linewidth}
		\centering
		\includegraphics[width=0.65\linewidth]{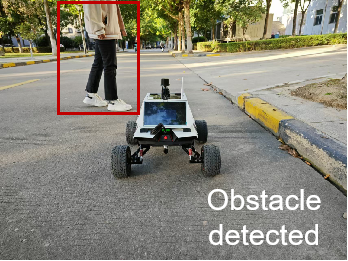}
            \caption{Pedestrian passes by}
		\label{fig:pedestrian passes by}%文中引用该图片代号
	\end{subfigure}
	% \centering
	\begin{subfigure}{0.23\linewidth}
		\centering
		\includegraphics[width=0.65\linewidth]{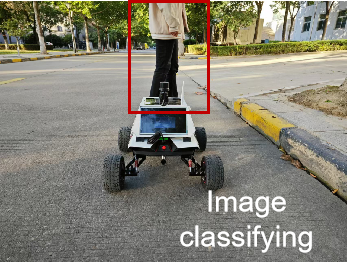}
            \caption{DL inference}
		\label{fig:dnn_infer}%文中引用该图片代号
	\end{subfigure}
        % \centering
	\begin{subfigure}{0.23\linewidth}
		\centering
		\includegraphics[width=0.65\linewidth]{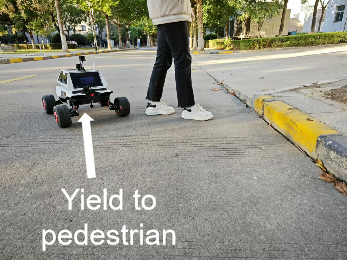}
            \caption{Vehicle avoids obstacles}
		\label{fig:avoid_obstacle}%文中引用该图片代号
	\end{subfigure}
        % \centering
	\begin{subfigure}{0.23\linewidth}
		\centering
		\includegraphics[width=0.65\linewidth]{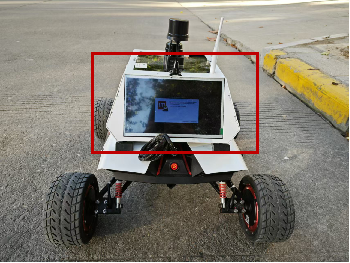}
            \caption{NVIDIA Jetson NX platform}
		\label{fig:case_platform}%文中引用该图片代号
	\end{subfigure}
        \vspace{-2mm}
	\caption{Case study of the DL-based image classification and obstacle avoidance.}
        \vspace{-2mm}
	\label{fig:case_example}
\end{figure*}

\begin{figure*}[t]
\centerline{\includegraphics[width=0.75\textwidth]{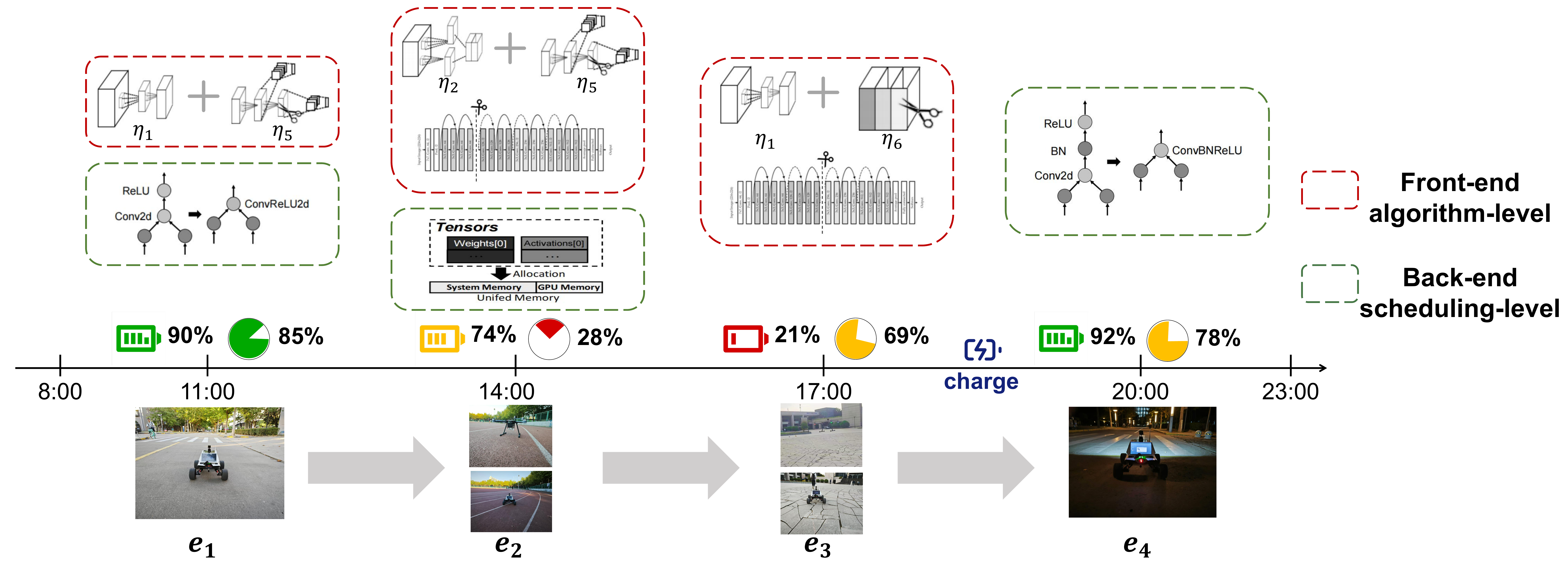}}
\vspace{-1mm}
\caption{Context-adaptive dynamic DL deployment using \rev{CrowdHMTware} under time-varying scenes.}
\vspace{-3mm}
\label{fig:case_study}
\end{figure*}

\subsection{Real-world Case Study}

We deploy \rev{CrowdHMTware} on a R300 vehicle equipped with NVIDIA Jetson Xavier NX and a P600 drone with NVIDIA Jetson Xavier NX to implement object classification and obstacle avoidance in a campus environment. 
The vehicle performs ground object classification (pedestrians, bicycles, cars), while the drone focuses on large-area objects (buildings, green spaces, birds). 
We test recognition tasks for a full day, facing diverse lighting conditions—strong contrasts during daylight and rapid changes in the evening—introducing challenges to classification accuracy.
Dynamic scene changes (\eg moving pedestrians and vehicles) were introduced to increase task complexity. 

As shown in \figref{fig:case_example}, when a pedestrian approaches, the vehicle uses \rev{CrowdHMTware}’s optimal cross-level deployment strategy for classification and obstacle avoidance.
To simulate real-world conditions, we increase the complexity of tasks, creating resource competition for memory and computation on both the vehicle and drone. 
The battery naturally depleted during runtime, as shown in \figref{fig:case_study}, with levels dropping from 90\% to 21\%, dynamically influencing the energy budget.

\rev{CrowdHMTware} continuously adapts cross-level strategies of DL deployment to balance performance and resources as scenes evolve throughout the day. 
For example, when the vehicle’s battery is at 90\% and memory at 85\% ($e_1$), \rev{CrowdHMTware} optimizes with elastic inference ($\eta_1$ + $\eta_5$) and operator fusion, focusing on real-time performance. 
As memory drops to 28\% ($e_2$), the system shifts to a lighter strategy, offloading tasks to the drone. 
With battery levels at 21\% ($e_3$), it prioritizes energy conservation, using elastic inference ($\eta_1$ + $\eta_6$) and offloading.
In the evening, when lighting variation affects classification, \rev{CrowdHMTware} adaptively adjusts model precision and adopts operator fusion to guarantee accuracy and responsiveness. 
\section{Related Work}
\label{sec:related}

\subsection{DL Model Specification Algorithms}

Previous studies have explored diverse DL model specialization methods to balance inference accuracy, latency, energy cost, and memory usage.

\textit{\textbf{Offline Specification}} method includes three types:
i) \textit{Hand-crafted DL model compression} like weight pruning~\cite{gou2021knowledge} rely on manual design to simplify model complexity but often fail to meet diverse performance needs. 
ii) \textit{On-demand model compression} such as DeepX~\cite{lane2016deepx}, AMC~\cite{he2018amc}, and AdaDeep~\cite{liu2020adadeep} uses trainable meta-learners to tailor compression strategies to different platforms, requiring online retraining that introduces substantial overhead, making them unsuitable for latency-sensitive applications. 
iii) \textit{One-shot neural architecture search (NAS)} automates the search for optimal architectures~\cite{fang2020fast}.
Yet they involve high overhead in scanning extensive candidate spaces and cannot operate online locally.

\textit{\textbf{Online Specification}} scheme includes three categories:
i) \textit{Supernetwork} encompasses a broad range of subnetworks for dynamic selection based on resource constraints, adapting to varying computational demands. 
Examples include Efficient Neural Architecture Search (ENAS)~\cite{pham2018efficient} which shares parameters to reduce training costs, Differentiable Architecture Search (DARTS)~\cite{liu2018darts} that optimizes network structure via gradient descent, and Neural Architecture Optimization (NAO)~\cite{luo2018neural} which uses algorithms to fine-tune designs, enhancing precision and efficiency.
ii) \textit{Elastic Model} selectively engages DL modules during inference to minimize complexity and energy use. 
Developments like Deep Elastic Networks~\cite{ahn2019deep} and Elastic Graph Neural Networks (ElasticGNN)~\cite{liu2021elastic} optimize performance and resource management by dynamically adjusting computational loads based on available resources.
iii) \textit{Adaptively  scalable} inference methods include NestDNN~\cite{fang2018nestdnn}, LegoDNN~\cite{han2021legodnn}, and FlexDNN~\cite{fang2020flexdnn}. 
NestDNN, as a model-grained DNN scaling method, suffers from significant performance degradation and accuracy loss due to the large gaps between models when switching at runtime.
While FlexDNN optimizes memory usage, it increases inference latency, making it difficult to meet diverse optimization goals in real-world scenarios.
Both NestDNN and LegoDNN are constrained to supporting CNN networks in mobile vision systems because they rely on filter pruning techniques for block/model compression.
All of these methods focus on optimizing individual levels, such as model architecture or memory management, which makes it difficult to balance various system metrics and achieve globally optimal results in systems where models and compilation engines interact.

\textit{\textbf{Multi-branch Model}} features several parallel paths, each processing data through distinct modules for efficient task handling. 
It simplifies training and enhances performance by offering a favorable optimization landscape. 
Recent advancements include an adaptive device-edge co-inference framework that uses multi-branch models~\cite{zhang2018deep} and a soft actor-critic method to optimize computational task distribution, improving resource usage and inference speed~\cite{niu2022adaptive}.
\sysname middleware can be applied on top of these online techniques to automate their selection and combination.

\subsection{DL Model Partition and Offloading}
In addition to DL model structure scaling, partitioning and offloading DL models to other devices can further aggregate available resources and enhance processing speed. 
Existing research primarily falls into two categories: 
\textit{Layer-level serial partitioning}:
Neurosurgeon~\cite{kang2017neurosurgeon} identifies cloud-only processing limitations and proposes fine-grained layer-level DL model partitioning by analyzing model structures and operator dependencies.
\textit{Intra-layer fine-grained parallel partitioning}:
For example, Band~\cite{jeong2022band} coordinates DL model workloads on heterogeneous processors by dynamically selecting scheduling schemes based on subgraph partitions and operator dependencies. 
Lu \etal~\cite{lu2019collaborative} supports various DL models with flexible fine-grained scheduling. 
However, they are constrained by the need for predefined input analysis and deep model layer slicing, reducing flexibility in dynamic offloading adjustments. 
\sysname integrates them to achieve hierarchical granularity, and decouples DL model pre-partitioning and offloading.

\subsection{Mobile DL Compilation Engine}
The engine for resource-efficient DL deployment on mobile and embedded devices aims to enhance memory scheduling flexibility, optimize computational efficiency, and adapt to heterogeneous devices. 
There are primarily two types of engines: \textit{interpreted} and \textit{compiled}.
\textit{Interpreted engines} interpret and parse model files, transforming DL models into efficient formats for inference or training and sequentially executing model operators. 
TFlite~\cite{tensorflowlite} is a lightweight engine tailored for mobile DL inference, offering core and custom operator support and featuring an optimized interpreter that minimizes load and execution latency. 
CMix-NN~\cite{capotondi2020cmix}, designed for inference on microcontrollers, focuses on model compression and supports diverse quantization strategies for layers, channels, and activations.
\textit{Compiled engines} convert models into machine code directly executable by hardware like CPUs and GPUs. 
TVM~\cite{chen2018tvm} introduces a comprehensive optimization stack, optimizing operator execution at the computation graph level, enhancing performance across various devices. 
Framework-specific optimizations include TensorFlow XLA~\cite{tensorflowxla} and TensorFlow Runtime (TFRT)~\cite{tensorflowruntime}, reducing redundant computing kernels, while TFRT optimizes DL inference across hardware for scalability, employing specific primitives for efficient execution.
Despite their capabilities, these engines often rely on fixed optimization strategies that may not perform optimally as device architectures and runtime conditions change. 
\sysname addresses this by enabling adaptive cross-level compilation optimization through dynamic context awareness, improving the effectiveness of DL deployments across mobile systems.

\subsection{Middlewares for Mobile DL Deployment}

Existing middleware often serves diverse domain like ~\cite{bode2023systematic}, blockchain~\cite{nasirifard2023orderlesschain}, and DL deployment~\cite{mendula2024furcifer}, with most DL deployment middlewares focusing on single-chip performance, often overlooking disparities between different chip types. 
For instance, Shao \etal ~\cite{shao2022novel} introduce a mobile virtual computing middleware leveraging CPU, GPU, and DSP for hierarchical model segmentation and scheduling. 
DNNTune~\cite{xia2019dnntune} is another middleware that enables layer-wise behavior analysis to optimize deployment strategies for mobile-cloud computing. 
CrowdHMTware, designed specifically for heterogeneous mobile devices, bridges the user performance needs with DL model deployment by appropriately mapping operators to resources, enhancing performance across varied devices.
% For latency and energy optimization, fine-grained, layer-level partitioning proves beneficial. 
% Neurosurgeon~\cite{kang2017neurosurgeon} automatically partitions DL model between mobile devices and data centers at the layer level, while SwapNet~\cite{wang2024swapnet} addresses delays from memory swapping on edge devices by optimizing memory operations during block swapping for deep learning tasks. 
Meanwhile, Furcifer~\cite{mendula2024furcifer}, a context-adaptive middleware for object detection, utilizes a container-based approach with low-complexity predictors and an optimized split DNN model for effective task offloading, reducing latency and enhancing performance.
Despite these advancements, existing middleware solutions often focus on stand-alone levels like model structure or compilation, restricting overall performance. 
CrowdHMTware, in contrast, aims to dynamically adjust cross-level optimization at runtime to adapt to changing contexts on mobile devices.

\section{Conclusion}
\label{sec:conclude}

This paper introduces CrowdHMTware, a dynamic context-adaptive DL deployment middleware for heterogeneous mobile devices. 
It overcomes performance bottlenecks associated with solely algorithmic or system scheduling through algorithm and system co-design. 
Specifically, \sysname comprises three key functional components: an elastic DL inference component that allows for retraining-free model structure scaling, a scalable DL offloading component that enables on-demand model offloading through decoupled multi-granularity model pre-partitioning and combination, and a model-adaptive back-end compilation engine that optimizes operator and resource scheduling. 
\sysname is implemented as a middleware system to obtain an automated adaptation loop. 
With experiments under heterogeneous and dynamic mobile deployment conditions, \sysname improves accuracy by up to 3.9\%, reduces latency by 10.3$\times$, and decreases energy consumption by 6.9$\times$ compared to existing baselines.

\section*{Acknowledgments}
This work was partially supported by the National Science Fund for Distinguished Young Scholars (62025205) the National Natural Science Foundation of China (No.62032020, 62472354, 62102317).

% \begin{acks}
% \end{acks}

\bibliography{hmt-base}
\bibliographystyle{IEEEtran}

\section{Biography Section}
\vspace{-10mm}
\begin{IEEEbiography}[{\includegraphics[height=1.25in,clip,keepaspectratio]{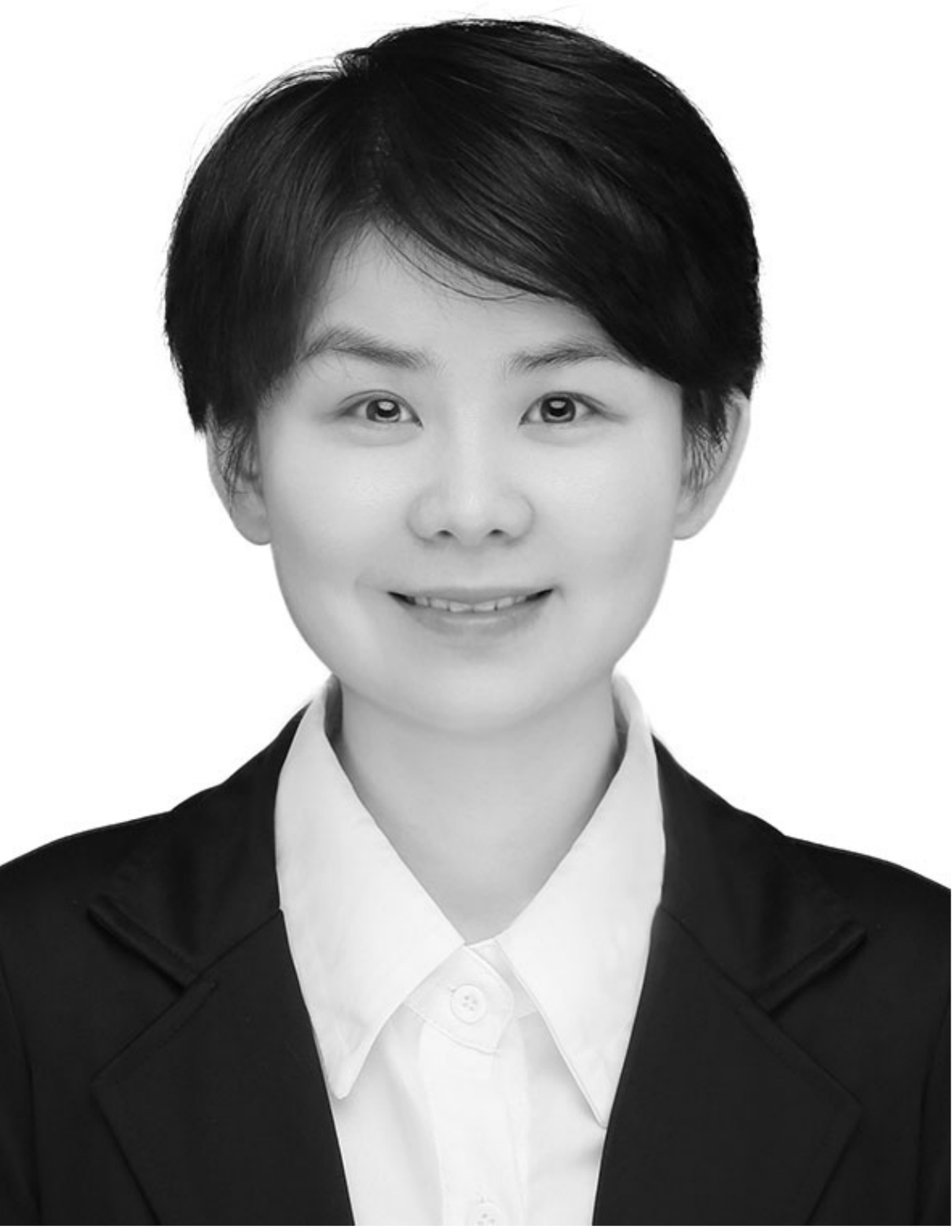}}]{Sicong Liu}
 received the PhD degree in school of computer science and technology from Xidian University in 2020. Now she is an associate professor with school of computer science, Northwestern Polytechnical University. Her research interests include mobile computing and resource-efficient mobile deep learning. She has served as the TPC member of MobiSys 2021 and BigCom 2021.
\end{IEEEbiography}
\vspace{-7mm}
\begin{IEEEbiography}[{\includegraphics[height=1.25in,clip,keepaspectratio]{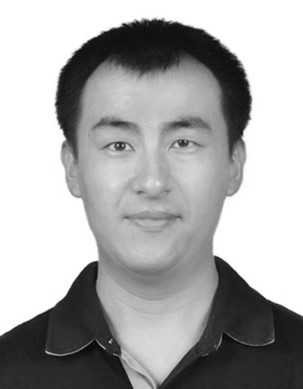}}]{Bin Guo}
 received the PhD degree in computer science from Keio University, Japan, and then was a postdoc researcher with Institut Telecom SudParis, France. He is a professor with Northwestern Polytechnical University, China. His research interests include ubiquitous computing, mobile crowd sensing, and HCI. He has served as an associate editor of the IEEE Communications Magazine and the IEEE Transactions on Human-Machine-Systems, the guest editor of the ACM Transactions on Intelligent Systems.
\end{IEEEbiography}
\vspace{-7mm}
\begin{IEEEbiography}[{\includegraphics[height=1.25in,clip,keepaspectratio]{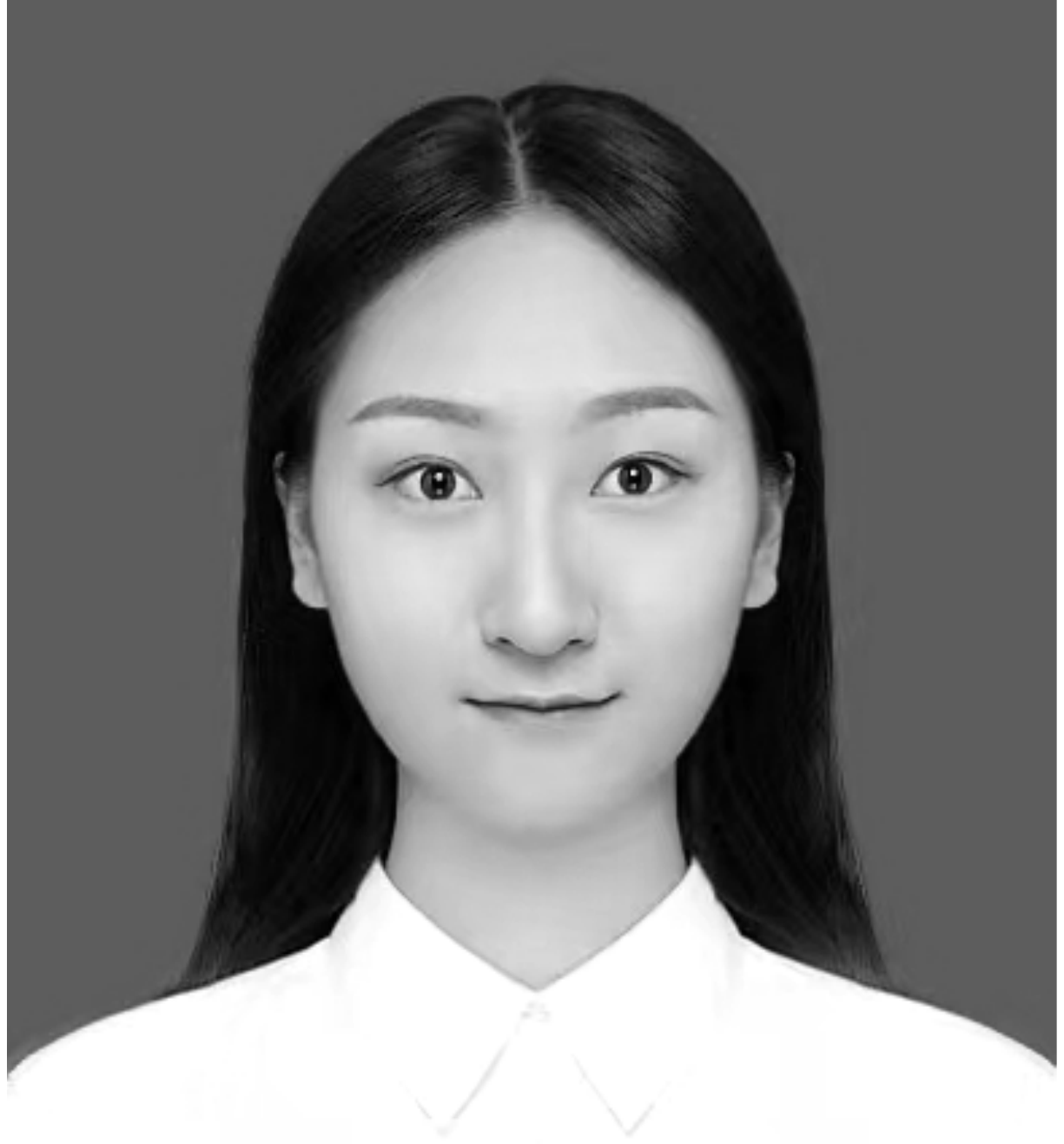}}]{Shiyan Luo}
is a Master's student at Northwestern Polytechnical University, China. She received her B.E. from Northwestern Polytechnical University in 2023. Her research interests are Mobile Computing and AIoT Systems.
\end{IEEEbiography}
\vspace{-7mm}
\begin{IEEEbiography}[{\includegraphics[height=1.25in,clip,keepaspectratio]{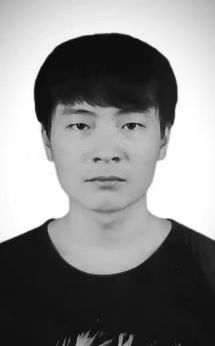}}]{Yuzhan Wang}
 received the B.E. degree from Northwestern Ploytechnical University, Xi'an, China, in 2022, where he is currently pursuing the Master degree. His current research interests include deep learning in resource-constrained devices and dynamic adaptive inference on mobile devices.
\end{IEEEbiography}
\vspace{-7mm}
\begin{IEEEbiography}[{\includegraphics[height=1.25in,clip,keepaspectratio]{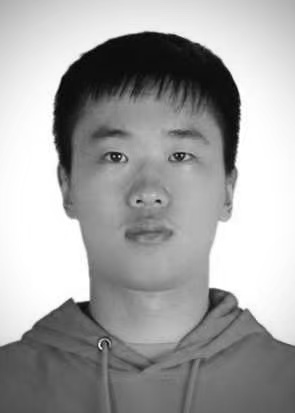}}]{Hao Luo}
is a Master's student at Northwestern Polytechnical University, China. He received his B.E. from Northwestern Polytechnical University in 2022. His research interests include mobile computing, model compression, and distributed model computing.
\end{IEEEbiography}
\vspace{-7mm}
\begin{IEEEbiography}[{\includegraphics[height=1.25in,clip,keepaspectratio]{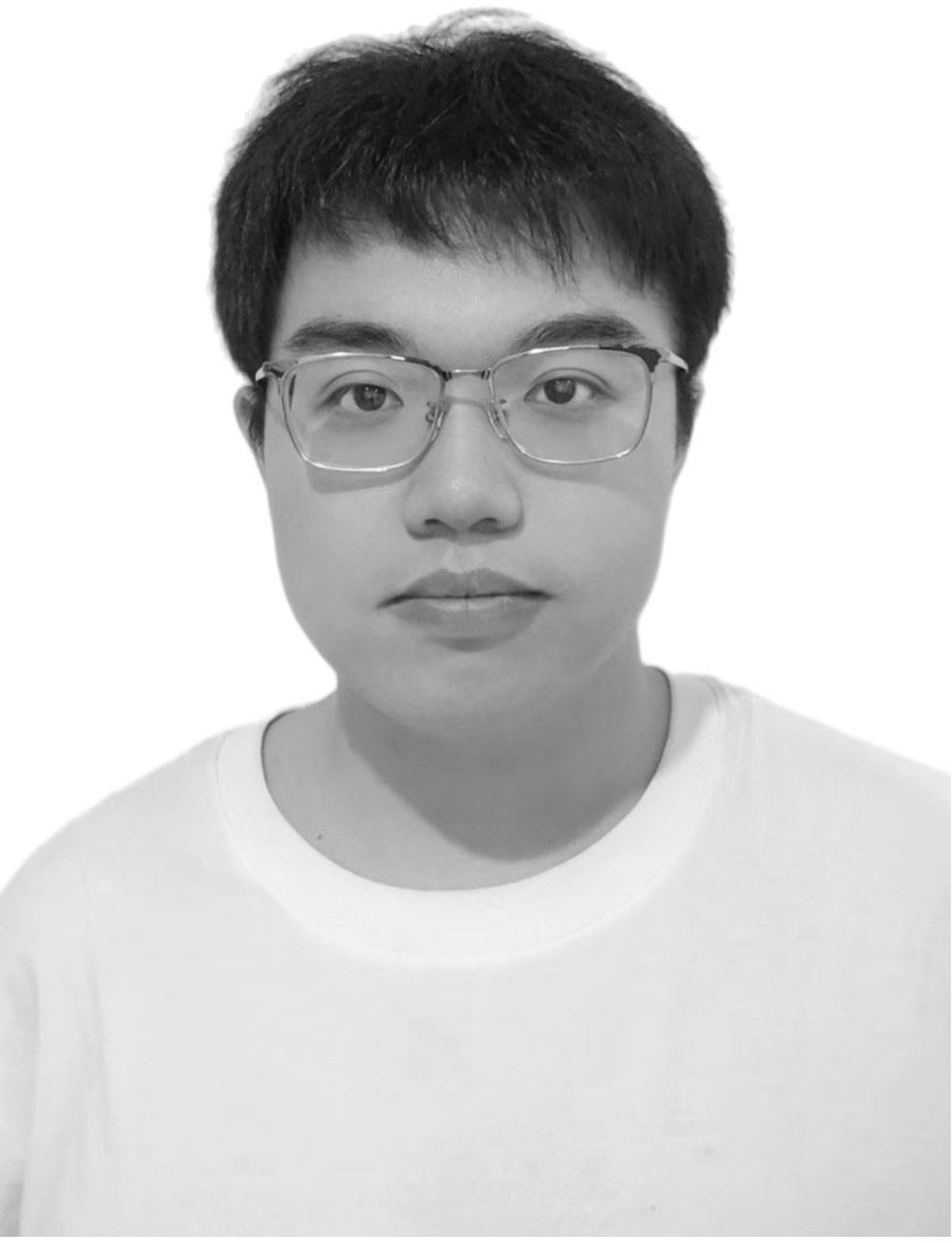}}]{Cheng Fang}
 received the B.E. degree from Northwestern Ploytechnical University, Xi'an, China, in 2022, where he is currently pursuing the Master degree. His current research interests include deep learning in resource-constrained devices and model adaptation on mobile devices
\end{IEEEbiography}
\vspace{-7mm}
\begin{IEEEbiography}[{\includegraphics[height=1.25in,clip,keepaspectratio]{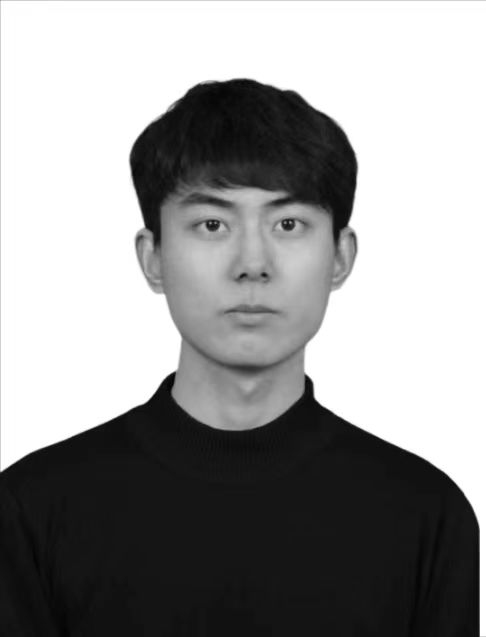}}]{Yuan Xu}
is a Master's student at Northwestern Polytechnical University, China. He received his B.E. from Northwest University in 2023. His interest is in Artificial Intelligence of Things (AIoT).
\end{IEEEbiography}
\vspace{-7mm}
\begin{IEEEbiography}[{\includegraphics[height=1.25in,clip,keepaspectratio]{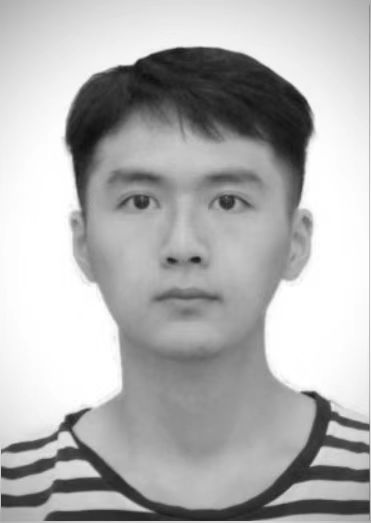}}]{Ke Ma}
received his bachelor degree in 2021, Computer Science, Northwestern Polytechnical University. He is now pursuing a phd degree in Computer Science, Northwestern Polytechnical University. His research interests are efficient AI on resource-constrained devices and deep model adaptation in ubiquitous environments.
\end{IEEEbiography}
\vspace{-7mm}
\begin{IEEEbiography}[{\includegraphics[height=1.25in,clip,keepaspectratio]{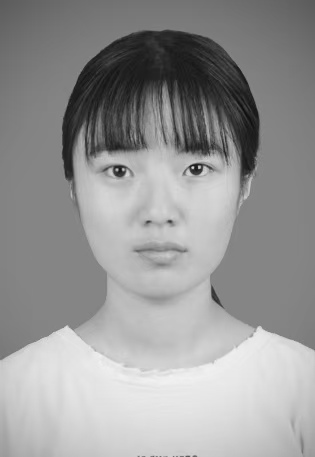}}]{Yao Li}
Yao Li received the B.E. degree from Northwestern Ploytechnical University, Xi'an, China, in 2023, where she is currently pursuing the Master degree. Her research interests include Mobile Computing, Federated Learning and Class Incremental Learning.
\end{IEEEbiography}
\vspace{-7mm}
\begin{IEEEbiography}[{\includegraphics[height=1.25in,clip,keepaspectratio]{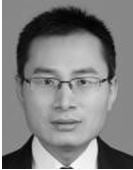}}]{Zhiwen Yu}
received the Ph.D. degree in computer science from Northwestern Polytechnical University, Xi’an, China, in 2005. He is currently the Vice President of Harbin Engineering University and a Professor at the School of Computer Science, Northwestern Polytechnical University. 
He was an Alexander Von Humboldt Fellow with Mannheim University, Germany, and a Research Fellow with Kyoto University, Kyoto, Japan. 
His research interests include ubiquitous computing, HCI, and mobile sensing and computing.
\end{IEEEbiography}
\vfill

\end{document}